\documentclass{article}
\usepackage{amd}
\usepackage[utf8]{inputenc}
\usepackage[T1]{fontenc}
\usepackage{booktabs}
\usepackage{amsfonts}
\usepackage{amsmath}
\usepackage{amssymb}
\usepackage{amsthm}
\usepackage{nicefrac}
\usepackage{microtype}
\usepackage{courier}
\usepackage{wrapfig}
\usepackage{tikz}
\usetikzlibrary{arrows.meta, positioning, fit, calc, decorations.pathreplacing}
\usepackage{placeins}
\usepackage{float}
\usepackage{hyperref}
\hypersetup{hidelinks}
\newif\ifmlsysreview
\mlsysreviewfalse
\newcommand{\tmark}{\texttrademark{}}

\IfFileExists{assets/AMD_logo.png}{\newcommand{\AMDlogo}{\includegraphics[height=0.28in]{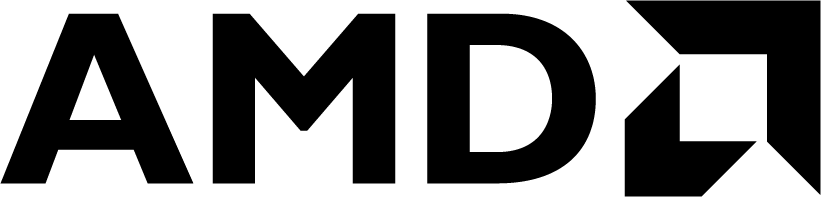}}}{}
\newcommand{\workshopnote}{NeurIPS 2026 Workshop on AXIOM: Foundations of Efficient Deep Learning.\par\noindent
NeurIPS 2026 Workshop on On-Device Intelligence: Foundation Models under Real-World Constraints.}
\makeatletter
\newcommand{\plainfootnote}[1]{\begingroup\let\thefootnote\relax
  \long\def\@makefntext##1{\noindent##1}\footnotetext{#1}\endgroup}
\makeatother

\icmlsetsymbol{intern}{$\dagger$}
\begin{document}
\onecolumn
\amdtitlerunning{The Price of Precision}
\amdtitle{The Price of Precision: A Calibrated Cost Model for Mixed-Precision LLM Serving}

\begin{amdauthorlist}
  \amdauthor{Mukund Athmaram}{amd,cornell,cor}
  \amdauthor{Manoj Rao}{amd}
  \amdauthor{Liang Ma}{amd}
\end{amdauthorlist}
\amdaffiliation{amd}{AMD}
\amdaffiliation{cornell}{Cornell University}
\amdcorrespondingauthor{Mukund Athmaram}{ma2388@cornell.edu}
\amdkeywords{mixed-precision quantization, LLM serving, cost model, mixture of experts, MXFP4, FP8}
\vskip 0.3in
\amdPrintAffiliationsInline
\plainfootnote{\workshopnote}

\begin{abstract}
When serving a large language model (LLM), it is standard practice to quantize the model to improve speed and
reduce memory use. Mixed precision assigns different precisions to different layers, because some layers are more
sensitive to quantization than others, and is usually chosen to preserve accuracy within a memory budget. But
quantizing a layer also changes how fast it runs, so every layer has both an ``accuracy impact'' and a ``speed
impact,'' and the goal is to choose, for each layer, the precision that best balances the two, often within a memory,
speed, or accuracy budget. Many existing allocators estimate the accuracy impact of each layer well, but for the speed
impact they rely on proxies such as the layer's size. We argue that the speed impact is more nuanced: it can be
limited by compute, weight loading, or key--value (KV) cache access, depending on the phase, batch size, and context
length, and it also depends on the hardware and on the kernel that runs each format. Across three hardware classes,
we find that 375 of 1{,}242 measured pairs of configurations ran slower with fewer bytes. To address
this, we build a cost model, calibrated once per model, device, and serving stack, that predicts the latency of
any configuration across operating points and anticipates these slowdowns, which byte-based estimates
cannot. We then use it in place of the byte-based time cost inside existing allocators, keeping their accuracy
measurements unchanged. Where bytes misjudge speed, the allocators select
configurations with up to $52\%$ lower latency. Finally, because the cost model's latency estimates change
with the operating point, the allocators choose different configurations at different operating points, and we measure how much is gained by changing precision as the device, batch size, phase, and
context change.
\end{abstract}

\section{Introduction}
Post-training quantization is standard for LLM serving. Mixed-precision schemes assign different bit-widths
to different weight tensors because tensors differ in both their sensitivity to quantization and their size.
Compressing a large tensor can save more memory, while keeping a sensitive tensor at higher precision can
preserve accuracy. Changing precision therefore changes both the tensor's accuracy impact and its
contribution to serving time. An allocator must choose these precisions together to find a good
accuracy--latency tradeoff for the whole model.

This is difficult for two reasons. First, we can measure the accuracy impact of quantizing one tensor,
but combining those measurements does not fully capture interactions when several tensors are quantized.
Second, a tensor's runtime depends on more than its size. Compute, weight loading, and precision-dependent
overhead contribute differently across batch sizes and between prefill and decode. Storing weights in four
bits rather than eight roughly halves their storage, so it is natural to assume that this also makes execution
faster. However, this is not always the case, as some lower-bit formats can be slower at a given operating point.

Many allocators nevertheless price time with a proxy for size. They pair a measure of each tensor's accuracy
impact, such as Hessian eigenvalues or traces~\citep{hawq,hawqv3}, learned assignments~\citep{dmx}, or variance
indicators~\citep{llmpq}, with a cost based on model size~\citep{hawq} or average bit-width~\citep{dmx}. Others
measure latency on the target device, with per-layer tables~\citep{hawqv3}, simulators~\citep{haq},
per-precision regressions~\citep{llmpq}, or profiled kernel tiles~\citep{mxmoe}. These must be re-profiled for each
stack and operating range, and regressions on FLOPs and bytes cannot see when a lower-precision format runs slower.

Size often tracks time, but not always, because of how the hardware runs each format. Graphics processing units (GPUs) overlap
compute with weight loading, so the time of a matrix multiplication is limited by either the compute rate or the
memory bandwidth, and the effect of quantization depends on which one is the limit. In decode, each step processes
only a few tokens, so most of the time goes to loading weights, and fewer bits mean less to load. The saving shrinks
when the kernel cannot unpack the low-bit values and their scales as fast as it loads them. In prefill, each
weight is reused across many tokens, so arithmetic dominates, and fewer bits help only if the hardware computes
faster in that format.

Converting between formats also takes time. QServe~\citep{qserve} reports 20--90\% runtime overhead from
these conversions in existing 4-bit integer methods on GPUs. In W4A16 (4-bit weights, 16-bit
activations), the multiply runs on FP16 tensor cores, so the overhead comes from dequantizing every 4-bit weight to
FP16 first. In W4A4 (4-bit weights and activations), the multiply runs in INT4, but the INT32 partial sum of every
quantization group must be converted to floating point to apply the group's scales, and this runs on the slower
general-purpose CUDA cores. Block-scaled formats
such as MX apply their scales inside the matrix instruction, which then runs at the rate of its wider input: on the
MI355X, 4-bit weights with 8-bit activations compute at the FP8 rate. Lowering the precision of the 
activations as
well means quantizing them at every step, often in a separate kernel whose fixed cost matters when each call has
little work. Whether a lower-precision format saves time therefore depends on the operating point, the hardware,
and the kernel, and has to be measured on the stack that runs it.

\begin{figure*}[!t]
\centering
\includegraphics[width=\textwidth]{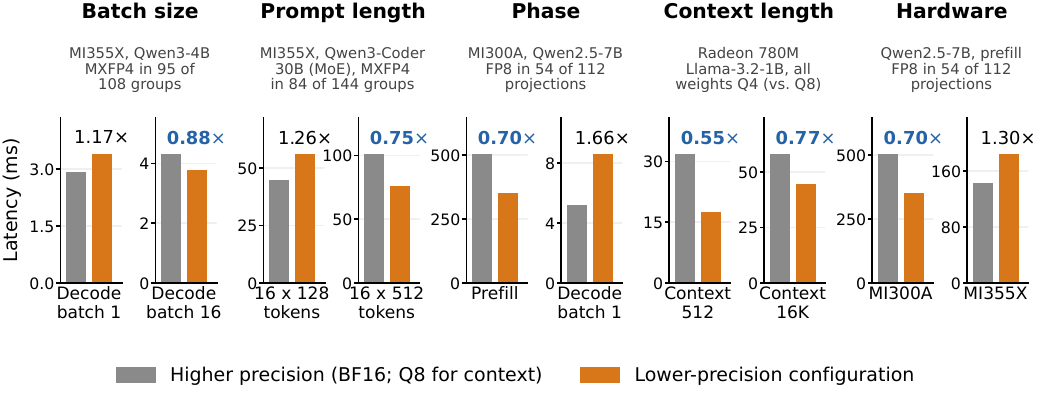}
\caption{Latency of one lower-precision configuration and its higher-precision reference, changing one condition per
panel. The reference is BF16, or Q8 in the context-length panel. Labels give the ratio to the reference, and blue marks where the lower-precision configuration is faster. Devices:
AMD Instinct\tmark{} MI355X GPU, AMD Instinct\tmark{} MI300A datacenter APU, and AMD
Radeon\tmark{} 780M Graphics. Appendix~\ref{app:fig1configs} shows each configuration.}
\label{fig:teaser}
\end{figure*}

Figure~\ref{fig:teaser} shows how much the speed impact of one precision configuration can change when a single
condition changes. The same configuration can be slower than bfloat16 (BF16) at batch~1 and faster at a larger batch, slower
on short prompts and faster on long ones, and faster in prefill but slower in decode. A lower-precision format
can also save less time as context grows, because reading the KV cache, which costs the same for every weight
format, takes a growing share of each step. Finally, a configuration that is faster than BF16 on one device can be
slower on another. In every case, the configuration's size, and therefore any cost based on size or bytes, stays
the same.

To address this, we develop a cost model that tracks the latency of a mixed-precision configuration across these operating points
and hardware, and evaluate it in three ways. First, as a predictor: it estimates a mixed-precision configuration's serving latency,
which we compare with measurements. Second, as the time cost inside an allocator: the allocator keeps its own
accuracy measurements but uses our model, instead of bytes, to estimate how each precision choice changes latency,
and we compare the configurations it selects in each case. We do this for allocators derived from
HAWQ-V3~\citep{hawqv3} and LLM-PQ~\citep{llmpq} under memory,
speed, and accuracy budgets, generating configurations for each operating point. With the cost model in place, the
allocator can search a large space of configurations quickly, without running them on the hardware. Third, because the model's time
estimate depends on the operating point, we measure what is gained by changing the precision configuration as the
operating point changes: across hardware (a configuration per device), batch size and load (activation precision
that switches with batch size while weights stay fixed), phase (weight precision per phase when prefill and decode
run on separate GPUs), and context length (weight precision that drops as the KV cache grows). We evaluate dense and mixture-of-experts (MoE)
models on three hardware classes: a datacenter GPU, a datacenter APU, and a laptop GPU.

\noindent\textbf{Contributions.} (i) A cost model that accounts for whether compute or memory bandwidth limits each
layer, evaluated as a latency predictor on three hardware classes (Sections~\ref{sec:method}--\ref{sec:predict}).
(ii) A drop-in time cost for existing allocators under memory, speed, and accuracy budgets
(Section~\ref{sec:existing}). (iii) An evaluation of how much is gained by switching precision as the operating
point changes (Section~\ref{sec:opdep}).

\section{The cost model}
\label{sec:method}
Given a unit $u$ at precision $p$, we estimate its time at the current phase, batch size, and
sequence length. A unit is a weight tensor used by a matrix multiplication, or a group of tensors
whose execution must be modeled together. Its time is $T_\text{unit}(u,p)=T_\text{work}(u,p)+h(u,p)$:
the general matrix multiplication (GEMM) work plus overhead for that precision.

\paragraph{GEMM work.} An individual GEMM $g$ with $W_g$ weights and $M_g$ token rows performs approximately
$2W_gM_g$ operations and loads $W_gb(p)$ weight bytes. Here $b(p)$ is the effective number of bytes
per weight, including quantization metadata. For example, a 4-bit microscaling floating-point (MXFP4) block stores 32 four-bit weights
in 16 bytes plus a one-byte scale~\citep{ocpmx}, so $b(\mathrm{MXFP4})=17/32=0.53125$ bytes per weight.
Dividing the operation count by effective throughput $r_\text{eff}(g,p)$ and the weight bytes by
bandwidth $\mathrm{BW}(p)$ gives compute and loading times, respectively. Following the roofline
approximation~\citep{williams2009roofline} explained in Appendix~\ref{app:roofline}, work time is their maximum:
\begin{equation}
t(g,p)=\max\!\left(
\underbrace{\frac{2W_gM_g}{r_\text{eff}(g,p)}}_{\text{compute time}},\
\underbrace{\frac{W_gb(p)}{\mathrm{BW}(p)}}_{\text{weight-loading time}}\right).
\label{eq:work}
\end{equation}
Summing $t(g,p)$ over the GEMMs executed by unit $u$ gives its work time $T_\text{work}(u,p)$.

\paragraph{Compute rate and bandwidth per format.} We model both terms of Equation~\ref{eq:work} per format.
For the compute rate, peak matrix rates differ by format, and the rate a kernel actually reaches depends on its
implementation. For the bandwidth, the device's memory bandwidth is the same for every format, but the rate at which
each format's kernels read weights is not, because unpacking sub-byte values, loading block scales, and keeping enough
loads in flight at small batch sizes lower it differently per format. For example, in MoE decode on the MI355X,
our calibration finds MXFP4 weights read at about $0.7\times$ the rate of BF16 weights. We therefore calibrate a compute rate $r(p)$ and
a bandwidth $\mathrm{BW}(p)$ per format.

\paragraph{Token rows in dense and MoE projections.} For a dense model, it is easy to predict how many token rows
each projection processes, but for an MoE model it is harder, because the router sends each token to only some
experts. A dense projection over all new tokens processes $M_g=BS$
rows, where $B$ is batch size and $S$ is the number of new tokens per sequence. Projections evaluated
only at selected positions use their executed row counts. For MoE projections,
we assume each token independently selects $k$ distinct experts uniformly from $E$ experts.
Following the active-expert estimate also used by \citet{oncescu2025opportunistic}, we spread
the $BSk$ assignments evenly across the expected $E_+$ active experts to approximate the rows
per active expert:
\begin{equation}
E_+=E\big[1-(1-k/E)^{BS}\big],\qquad M_\text{expert}\approx\frac{BSk}{E_+}.
\label{eq:activeexperts}
\end{equation}
For an expert unit containing all experts in one layer, each $t(g,p)$ uses one expert's
projection weights and $M_g=M_\text{expert}$. We sum these times over one expert's projections
and multiply by $E_+$ to obtain $T_\text{work}(u,p)$. Appendix~\ref{app:roofline} analyzes the
validity and limitations of our balanced-routing approximation.

Unpacking or converting weights inside a GEMM is already captured by the rate and bandwidth we calibrate for its
format. The term $h(u,p)$ adds work outside the GEMM, such as a separate activation-quantization kernel or kernel
launches.

We sum unit times, attention time, and a shared baseline $f_\text{base}$, fitted per phase, for remaining work
that does not depend on precision, such as normalization, residual additions, and serving-engine overhead:
\begin{equation}
\begin{aligned}
T_\text{step}(\text{cfg};B,S,L)
 &=f_\text{base}+T_\text{attention}(B,S,L)\\
 &\quad+\sum_u T_\text{unit}\big(u,\text{cfg}[u]\big).
\end{aligned}
\label{eq:step}
\end{equation}
Attention time covers
scores, value aggregation, and KV-cache access.

For fixed model dimensions and cache precision, we use $T_\text{attention}=a_\text{prefill}BS^2$ for
initial prefill and $T_\text{attention}=a_\text{decode}BL$ for standard single-token decode, where $L$
is the attended context length, including the current token. Prefill query--key pairs grow
quadratically with $S$. Single-token decode adds one query per sequence that attends over $L$
cached keys and values, giving a modeled cost linear in $L$.
This assumes each step is either a prefill or a decode step. In serving systems that use continuous batching with
chunked prefill, one step can mix prompt chunks from some requests with decode tokens from others. The
precision-dependent terms still apply, since they depend only on how many token rows each weight multiplies, which is
now the step's prompt tokens plus its decode tokens. The baseline and attention terms are instead summed over the
prompt and decode tokens in the step, with pure prefill and pure decode as special cases
(Appendix~\ref{app:colocated}).

\subsection{Calibrating the model to a deployment}
\label{sec:calib}
For each model, device, and serving stack, calibration supplies (i) a compute rate $r(p)$ per format and the blend scale $M_{1/2}$
below, (ii) a bandwidth $\mathrm{BW}(p)$ per format, (iii) overheads $h(u,p)$, (iv) the per-phase baseline
$f_\text{base}$, and (v) the attention coefficients $a_\text{prefill},a_\text{decode}$.

\paragraph{Compute rate at small batches.} We calibrate each format's rate $r(p)$ with enough token rows to keep
its kernel busy. With only a few rows, the kernel runs closer to the reference rate, because many runtimes then
unpack the weights and multiply in the reference precision. To account for this, we interpolate between the
reference rate and the format rate as the number of token rows grows:
\begin{equation}
r_\text{eff}(g,p)=(1-\eta_g)r_\text{ref}+\eta_g r(p),\qquad
\eta_g=\frac{M_g}{M_g+M_{1/2}}.
\label{eq:blend}
\end{equation}
Here $r_\text{ref}$ is the calibrated compute throughput at the model's original precision, and the blend scale
$M_{1/2}$ is the row count where the two rates have equal weight.
On every platform, we jointly fit rates, bandwidths, this blend scale, and overheads, shared by prefill and
decode, together with phase-specific baseline and attention costs.
Calibration serves 7--8 configurations, including each uniform format, over 9--12 workloads per device, which takes
about 1--2 device-hours per model, device, and serving stack. The model is then frozen before any prediction or
selection is evaluated. Appendix~\ref{app:roofline} gives the procedure and fitted parameter values
(Table~\ref{tab:effbw_params}), and Appendix~\ref{app:calrobust} tests robustness to the calibration set.

\section{Predicting served speed}
\label{sec:predict}
\begin{figure*}[!t]
\centering
\includegraphics[width=\textwidth]{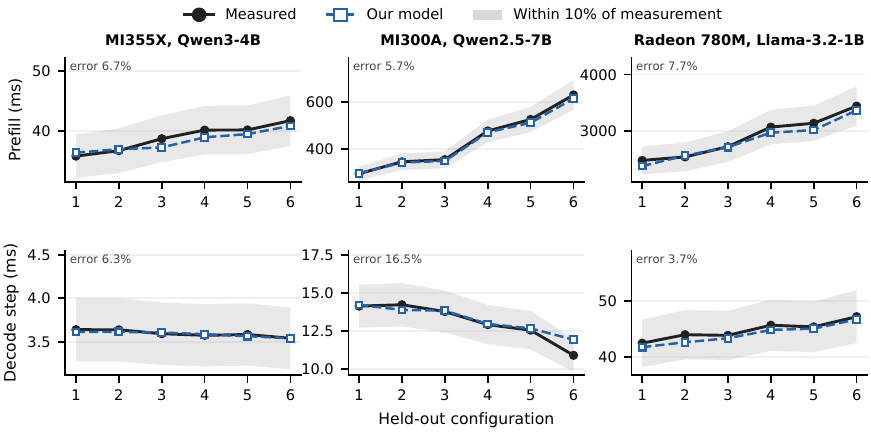}
\caption{Measured and predicted latency of six held-out configurations per device: prefill (top) and one decode step
(bottom), each averaged over six workloads. Each panel gives our model's mean error over individual configuration--workload cells.}
\label{fig:predmeas}
\end{figure*}

We evaluate Qwen3-4B~\citep{qwen3} on the AMD Instinct\tmark{} MI355X GPU (BF16/MXFP4),
Qwen2.5-7B-Instruct~\citep{qwen25instruct} on the AMD Instinct\tmark{} MI300A datacenter accelerated processing unit (APU) with BF16 and 8-bit floating-point (FP8) weights, and
Llama-3.2-1B-Instruct~\citep{llama32} on AMD Radeon\tmark{} 780M Graphics (F16/Q8). Figure~\ref{fig:predmeas}
compares predicted and measured mean prefill and decode latency on configurations
and workloads excluded from calibration (on the Radeon~780M, 4 of 12 phase-workloads share a setting with
calibration). On all three platforms, we sample six configurations by randomly quantizing
attention and feed-forward groups at the same six probabilities (Appendix~\ref{app:roofline}).

\begin{figure}[!t]
\centering
\includegraphics[width=0.62\linewidth]{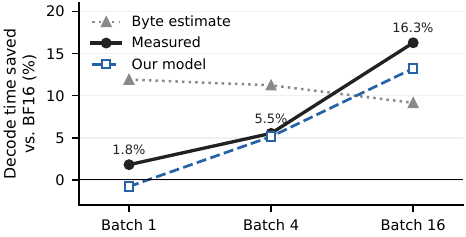}
\caption{Decode time saved by MXFP4 relative to BF16 at three batch sizes (Qwen3-Coder-30B-A3B on MI355X, mean of
three held-out configurations): measured, our model, and the byte estimate.}
\label{fig:bytetracking}
\end{figure}

Our model has lower mean error than byte estimates (Appendix Table~\ref{tab:proxies}) and, unlike bytes, captures
what changes the ranking. For example, in the MoE model in Figure~\ref{fig:bytetracking},
more experts are active at larger batches, so the time saved by quantizing grows with batch size. Our model counts
the active experts and also predicts a saving that grows with batch size, while the byte estimate predicts the same
time saved at every batch, which is a slightly smaller share of the longer step at larger batches.

\begin{figure}[!t]
\centering
\includegraphics[width=0.62\linewidth]{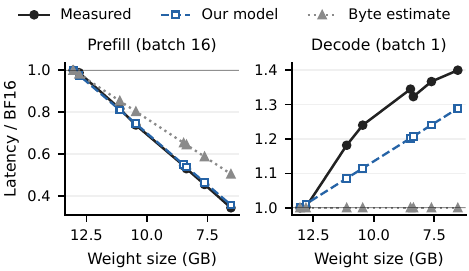}
\caption{Latency relative to BF16 of eight Qwen2.5-7B configurations on MI300A, ordered by weight size from all BF16 to
all FP8: measured, our model, and the byte estimate.}
\label{fig:fewerbytes}
\end{figure}

\paragraph{Fewer bytes being slower.} Figure~\ref{fig:fewerbytes} shows held-out configurations of Qwen2.5-7B on
the MI300A, from all BF16 to all FP8. As the weights shrink, prefill gets faster but batch-1 decode gets slower.
At batch~1 each layer multiplies a single token, so time goes to reading weights, and on this
stack the FP8 kernel reads them more slowly than the BF16 kernel and adds a separate step to convert activations to
FP8. In prefill, however, many tokens share each weight read, so FP8's faster arithmetic wins. A byte estimate can only
predict that fewer bytes are faster or equally fast, so it cannot represent this slowdown. Our model, with its
per-format bandwidth, follows it. This holds across hardware classes: roughly a third of measured pairs of configurations
ran more than $2\%$ slower with fewer bytes. Byte estimates cannot predict any of them,
and a fitted FLOP and byte regression~\citep{llmpq} almost none, while our model predicts 269 of the 375, and $81\%$ of
those that are consistent when we measure each configuration several times. When our model predicts a slowdown,
76\% of the time the configuration is measured to be slower. Figure~\ref{fig:c1} and
Table~\ref{tab:proxies} in the appendix give every platform and metric.

\section{Allocation with the cost model}
\label{sec:allocator}
We choose a supported precision $p_u\in\mathcal{P}_u$ for each unit $u$ under a latency or accuracy
budget. For each choice, we measure the accuracy-cost change and model the time change relative to
the reference precision $p_0(u)$.

\paragraph{Accuracy change.} We quantize one unit at a time, keeping the rest at reference precision.
Over WikiText~\citep{merity2017pointer}, the mean Kullback--Leibler (KL) divergence from reference
to quantized next-token distributions gives $A(u,p)$. Repeating across units and formats builds an accuracy
table for each model. Appendix~\ref{sec:analysis} shows an example. For a change from $p_0(u)$ to $p$,
\begin{equation}
\Delta A(u,p)=A(u,p)-A\big(u,p_0(u)\big).
\label{eq:daccuracy}
\end{equation}

\paragraph{Time change.} Using the cost model, we normalize the unit-time difference by the
reference configuration's predicted step time $T_0$ at the same operating point:
\begin{equation}
\Delta S(u,p)=\frac{T_\text{unit}(u,p)-T_\text{unit}\big(u,p_0(u)\big)}{T_0}.
\label{eq:dspeed}
\end{equation}

\paragraph{Choosing the configuration.} Each unit takes exactly one precision from its menu, so the choice is a
multiple-choice knapsack. We solve
\begin{equation}
\begin{aligned}
\min_{\{p_u\}}\quad&\sum_u\Delta A(u,p_u)\\
\text{s.t.}\quad&\sum_u\Delta S(u,p_u)\le\tau,\quad p_u\in\mathcal{P}_u.
\end{aligned}
\label{eq:knapsack}
\end{equation}
For latency budget $D$, $\tau=D/T_0-1$, and for a target of $x$ lower latency than the reference, $\tau=-x$.
With an accuracy budget instead, we solve
\begin{equation}
\begin{aligned}
\min_{\{p_u\}}\quad&\sum_u\Delta S(u,p_u)\\
\text{s.t.}\quad&\sum_u\Delta A(u,p_u)\le\epsilon,\quad p_u\in\mathcal{P}_u,
\end{aligned}
\label{eq:accuracybudget}
\end{equation}
where $\epsilon$ is the allowed increase in the additive accuracy cost. A dynamic program over a discretized time or
accuracy grid solves both problems exactly for this discretized additive model. With an accuracy budget, it selects
the fastest state within the budget (Algorithm~\ref{alg:allocate}, Appendix~\ref{app:opt}). Appendix~\ref{sec:opdivergence}
evaluates how well this additive approximation of KL holds.
We use this allocator in two ways. To compare time costs, we keep it fixed and swap only its time estimate, bytes
versus our model (Section~\ref{sec:existing}). To follow the operating point, we run it separately for each device,
phase, batch size, or context range and compare the resulting configurations (Section~\ref{sec:opdep}).

\section{Improving existing allocators}
\label{sec:existing}
\label{sec:swap}
The allocator above has an accuracy cost per unit (Equation~\ref{eq:daccuracy}), a time cost per unit
(Equation~\ref{eq:dspeed}), and a search under a budget. Existing mixed-precision methods mostly differ in how they
measure accuracy and how they search, and many use size or bytes for the time cost (Section~\ref{sec:related}). We
keep each allocator's accuracy measure and search fixed and only swap the time cost, so any difference in the
selected configurations comes from how time is estimated. We do this in our own allocator, which measures accuracy as the KL divergence of
quantizing each unit (Section~\ref{sec:allocator}). We also do it in two allocators
derived from published methods, each with its own sensitivity measure and solver (Appendix~\ref{app:realalloc}).
The one derived from HAWQ-V3~\citep{hawqv3} measures sensitivity with Hessian traces, and the one derived from
LLM-PQ~\citep{llmpq} uses its variance indicator.

Allocators are used under a memory, speed, or accuracy budget, and the time cost plays a different role in each.
With a memory budget, the allocator trades time against accuracy. With a speed budget, time is the limit it has to
stay under, and with an accuracy budget, time is what it minimizes. We test our cost under all three.

\subsection{Memory budget}
\label{sec:membudget}
\begin{figure}[!htb]
\centering
\includegraphics[width=0.62\linewidth]{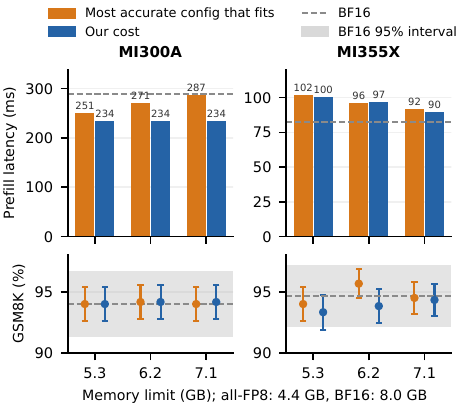}
\caption{Prefill latency (32 prompts of 256 tokens) and GSM8K accuracy at three memory limits (Qwen3-4B, BF16 or
block-scaled FP8 per layer), averaged over two sensitivity measures.}
\label{fig:membudget}
\end{figure}

With a memory budget, the allocator chooses among the configurations that fit in a given size. The baseline, derived
from HAWQ-V3, picks the most accurate configuration that fits. Ours uses the LLM-PQ-derived objective, which weighs
accuracy against estimated time, with our cost model giving the time. We run both allocators
with both sensitivity measures, Hessian traces and the variance indicator, at three memory limits between the all-FP8 size (4.4\,GB) and the BF16 size
(8.0\,GB). In prefill, the baseline quantizes the layers where FP8 costs the least accuracy, regardless of speed.
Ours quantizes the layers where FP8 saves the most time for the accuracy it costs. On MI300A, ours is 7--18\% faster
with measured accuracy within noise
(Figure~\ref{fig:membudget}).

\begin{figure*}[!t]
\centering
\includegraphics[width=\textwidth]{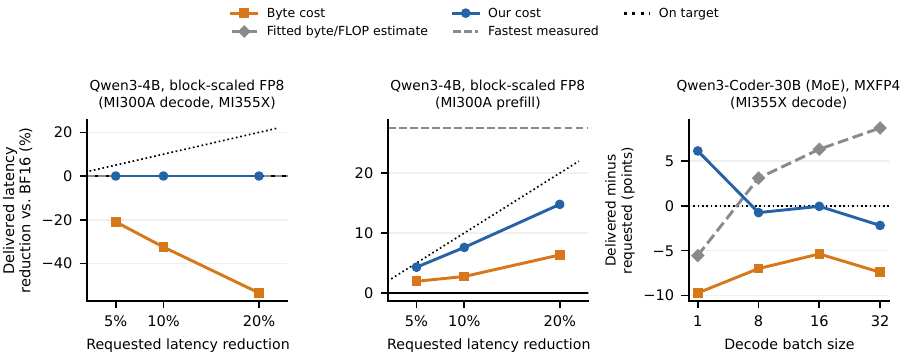}
\caption{Speed budgets. Left and middle: latency reduction delivered versus requested (Qwen3-4B, block-scaled FP8),
where points on the dotted line meet the target. Right: delivered minus requested reduction on a MoE model by decode
batch size, averaged over the targets, where 0 means on target.}
\label{fig:speedbudget}
\end{figure*}

\subsection{Speed budget}
\label{sec:speedbudget}
With a speed budget, the allocator returns the most accurate configuration it estimates will meet a latency target.
We ask both derived allocators, each with its own sensitivity measure, for configurations that are 5, 10, and 20\%
faster than BF16 at each operating point
(Figure~\ref{fig:speedbudget}). A byte cost assumes that 20\% fewer weight bytes gives a 20\% speedup. In decode, no
FP8 configuration is faster than BF16 with the stack's FP8 kernel, so the byte cost promises every target and then
misses it.
Our cost instead reports that these targets cannot be met and keeps BF16, the fastest configuration we measured
there. In prefill on MI300A, where FP8 is faster, our selections get closer to each target, and accuracy stays within
noise of BF16 throughout.

For the MoE model, with the per-unit KL accuracy costs, the time saved by quantizing experts depends on batch size, since few experts are active at
batch~1 and most are active at batch~32. Bytes cannot see this and miss in opposite directions as batch size changes.
Our cost counts active experts and stays close to the target at every batch size (Figure~\ref{fig:speedbudget},
right). Per-layer timings measured in isolation, which some allocators use~\citep{hawqv3}, also miss this, because a
layer timed alone does not behave like a served MoE step (Appendix~\ref{app:budgets}).

\begin{figure*}[!t]
\centering
\includegraphics[width=\textwidth]{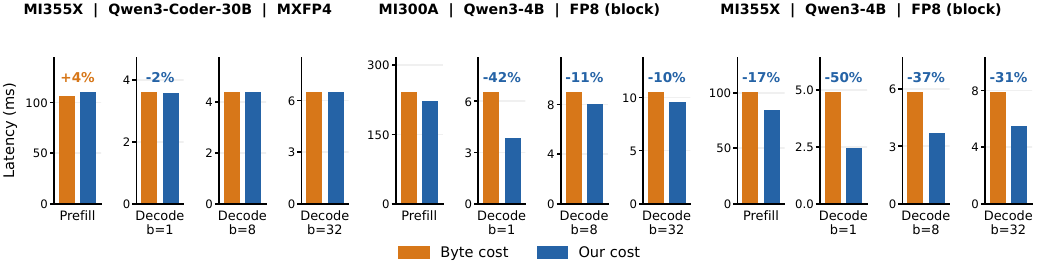}
\caption{Latency of the configurations our allocator selects with a byte cost or with our cost, under the same
accuracy budget. Labels mark differences outside the 95\% interval.}
\label{fig:c2bars}
\end{figure*}

\subsection{Accuracy budget}
\label{sec:accbudget}
With an accuracy budget, the allocator returns the fastest configuration it estimates is within an allowed accuracy
loss. For our allocator, the budget is on the additive KL proxy of Equation~\ref{eq:accuracybudget}, set to 10, 20, or 50\% of the proxy
loss of quantizing every layer, and we measure the accuracy of each selected configuration afterward. Where bytes
track speed, as with MXFP4, both costs select the same or nearly the same configurations, with one MoE prefill point
4\% slower (Figure~\ref{fig:c2bars}, left). Block-scaled FP8 is different. With the stack's kernel, it is slower than
BF16 in small-batch decode on MI300A and at every operating point on MI355X. A byte cost spends the budget on exactly
these layers, while ours keeps them in BF16 and uses FP8 only where it is faster (Figure~\ref{fig:c2bars}, middle and
right). Measured accuracy stays within noise between the two costs. Across models and GPUs, our selections have up to
52\% lower latency than the byte-cost selections (Appendix~\ref{app:full}). The allocators derived from HAWQ-V3 and
LLM-PQ, with their own sensitivity measures, behave the same way. With our cost, they select faster configurations at 77 of 96 points and slower ones at
none (Appendix~\ref{app:realalloc}).

\paragraph{Following a kernel change.} Our cost model also follows changes to the kernels. The FP8 slowdown on MI355X comes from the kernel configuration the stack
ships. With a tuned configuration, FP8 becomes faster than BF16 in decode. Once
recalibrated on the tuned stack, our model switches to FP8-heavy configurations, while the byte cost returns the same
configurations on both stacks (Appendix~\ref{app:kerneltuning}).

\section{Precision across operating points}
\label{sec:opdep}
A byte cost gives a configuration the same cost at every operating point, so an allocator using it returns one
configuration for all of them. Our model takes the operating point as input, which means the best configuration can
change as the operating point changes. In this section we measure how much is gained by actually following it.

One might ask why phase, batch size, and prompt length are separate parts of the operating point, since for the
linear layers they only change the number of token rows each weight multiplies. They still differ in how attention
grows, in how much accuracy low-precision activations cost, and in how precision can be switched, so each subsection
below changes one part of the operating point.

\subsection{Hardware}
\label{sec:ophw}
\begin{figure}[!htb]
\centering
\includegraphics[width=0.62\linewidth]{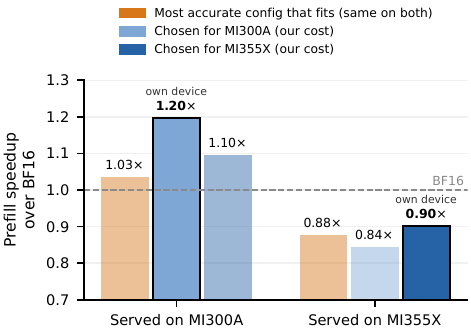}
\caption{Prefill speedup over BF16 of each device's configuration served on both devices, at the same memory limit
(Qwen3-4B, block-scaled FP8). BF16 does not fit the limit. On MI355X, block-scaled FP8 is slower than BF16 in
prefill, so the best choice there is the least slow.}
\label{fig:configsxhw}
\end{figure}

A cost that ignores the device returns the same configuration everywhere, while ours is calibrated per device and
gives each device its own. To see whether this matters, we fix the memory limit and serve each device's configuration
on both devices (Figure~\ref{fig:configsxhw}). Each configuration is fastest on the device it was chosen for. The
MI300A configuration uses FP8 widely, since its FP8 kernel is fast, while the MI355X configuration keeps more layers
in BF16, where FP8 is slow, and Qwen2.5-7B shows the same reversal (Appendix~\ref{app:full}). Accuracy, on the
other hand, transfers across devices, so only the time cost has to be recalibrated for new hardware.

\subsection{Batch size and load}
\label{sec:opbatch}
\begin{figure*}[!t]
\centering
\includegraphics[width=\textwidth]{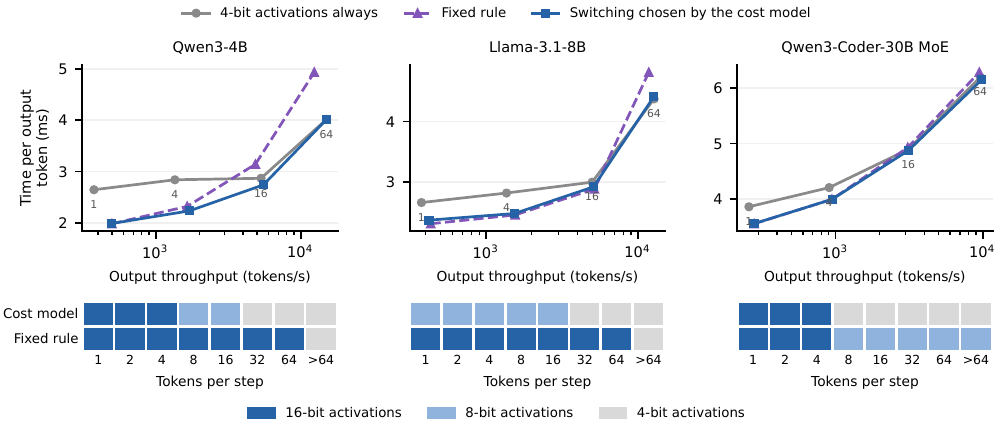}
\caption{Top: time per output token versus output throughput with MXFP4 weights on MI355X, one point per number of
concurrent requests. Bottom: activation precision used at each number of tokens per step, which in decode is the batch size.}
\label{fig:c3}
\end{figure*}

\paragraph{Weights or activations?} Should the configuration change with batch size? Switching weights per batch
saves little time, and it would need a second copy of every weight in memory or a reload that costs far more than a
decode step. Activation precision is much easier to switch, since the same MXFP4 weights can be multiplied by BF16,
FP8, or MXFP4 activations. Its cost also changes with batch size for the same reason as for weights
(Appendix Table~\ref{tab:moecost-full}). At small decode batches the multiplies wait on weight loading, so
higher-precision activations are nearly free, while at large batches they are compute-bound and expensive. Accuracy
goes the other way, since low-precision activations cost accuracy in decode but not in prefill (Appendix
Figure~\ref{fig:phasesens-full}), as also found by~\citet{disaggquant} and~\citet{mixquant}. So higher-precision
activations belong at small batches and low-precision ones at large batches, and the cost model decides where to
switch. Kernel libraries switch MoE activation precision at a fixed token threshold~\citep{aiter}, and prior work
switches it by phase~\citep{disaggquant} or between drafting and verification for speculative decoding~\citep{qspec}. Switching where the
cost model says is faster than static 4-bit activations at low load and matches them at high load
(Figure~\ref{fig:c3}), and at batch~1 it raises GSM8K accuracy over 4-bit activations by 3.6 and 2.6 points on
Qwen3-4B and Llama-3.1-8B
(Appendices~\ref{app:c3load} and~\ref{app:openloop}).

\subsection{Prefill and decode}
\label{sec:opphase}
\begin{figure*}[!t]
\centering
\includegraphics[width=0.85\textwidth]{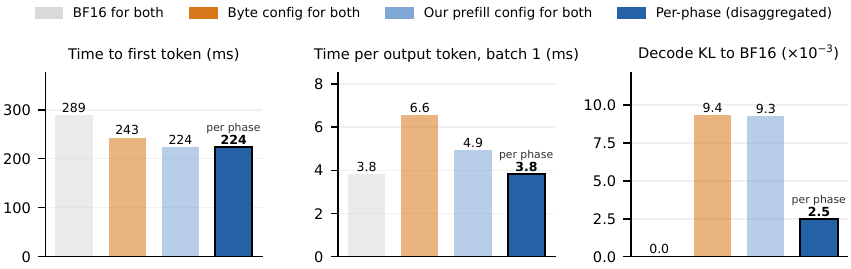}
\caption{Prefill and decode with a configuration per phase versus one configuration for both phases (MI300A,
Qwen3-4B, block-scaled FP8).}
\label{fig:disagg}
\end{figure*}

Prefill is compute-bound and decode is limited by weight loading, so the best configuration differs between them.
Serving systems that run prefill and decode on separate GPU pools~\citep{zhong2024distserve} can give each pool its own
precision. Mix-Quant and Disaggregated Quantization already quantize the two phases differently, with one format for
all layers in each phase~\citep{mixquant,disaggquant}. We instead choose each pool's configuration per layer with the
cost model. On MI300A, it picks an FP8 mix for prefill,
where FP8 is faster, and BF16 for decode, where its FP8 kernel is slow, and only this per-phase deployment is fast in
both phases (Figure~\ref{fig:disagg}). Since decode then runs at full precision, it also stays much closer to BF16 in
accuracy, even though it reads a KV cache written by the FP8 prefill (Appendix~\ref{app:disagg}). Accuracy differs by
phase too. On MI355X with MXFP4 on real disaggregated pools, quantizing only prefill costs far less accuracy than
quantizing decode, and the most sensitive layers differ, so giving each pool its own accuracy table yields
configurations closer to BF16 at the same speed (Appendix~\ref{app:disagg}).

\paragraph{Colocated prefill and decode.} Most deployments run both phases on the same GPUs, so one configuration has
to serve both. Here the allocator weights each configuration's cost by the mix of prefill and decode work in the
traffic. For example, for GSM8K traffic, which is about 95\% decode, it picks the decode configuration, which serves 12\% faster
per output token than the prefill configuration (Appendix~\ref{app:colocated}).

\subsection{Context length}
\label{sec:kv}
\label{sec:context}
As context grows, reading the KV cache takes a growing share of each decode step, and that time is the same for
every weight format. Our model captures this through the attention term and tracks decode latency through 16K tokens
on the Radeon~780M (Appendix Figure~\ref{fig:kvcontext}).

\paragraph{Cache precision.} The cache can be quantized too. Its access time scales roughly with $b(\kappa)L$, where
$b(\kappa)$ is bytes per cached element in format $\kappa$, so the allocator can treat cache precision as one more
choice. For each cache format, it subtracts the predicted attention time from the budget and allocates the weights
with the time that remains (Appendix~\ref{app:kvdegrad}). Cache precision also costs accuracy, so the cache shares the
accuracy budget with the weights.

\begin{figure*}[!t]
\centering
\includegraphics[width=\textwidth]{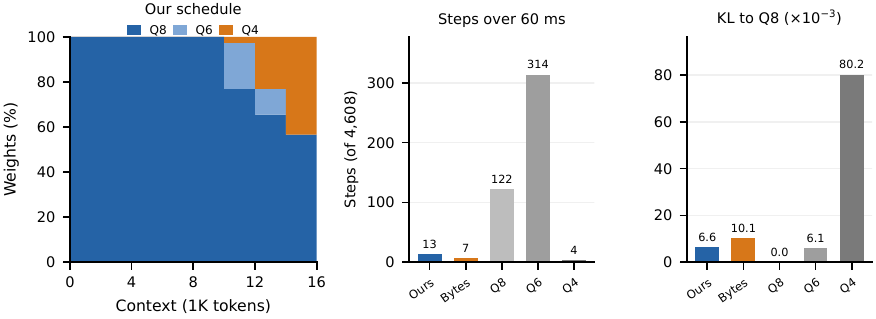}
\caption{Our weight schedule as context grows (Llama-3.2-1B, Radeon~780M), with steps over 60\,ms and KL to Q8 for
each policy.}
\label{fig:scheduleconfirm}
\end{figure*}

\paragraph{Scheduling weights as context grows.} Since growing context leaves less of the latency budget for the
weights, can we lower weight precision as context grows, and can our allocator decide when? PMPD~\citep{pmpd} lowers
whole-model precision during generation with a schedule tuned for accuracy. We instead schedule per layer by context
length to stay within a latency budget. Here we switch weights rather than activations, because context changes
attention and cache reads but not the token rows each weight multiplies, so an activation format's cost barely
changes with context. We run this on the Radeon~780M with a modified llama.cpp~\citep{llamacpp} that keeps every
precision resident, so a switch only changes which copy is used. The same schedules can also run without resident
copies (Appendix~\ref{app:noresident}). For each context range, the cost model picks the most accurate
configuration within the budget (Appendix~\ref{sec:hardware}). The schedule keeps Q8 through 10K tokens and then
moves layers to Q6 and Q4 (Figure~\ref{fig:scheduleconfirm}). With a schedule, decode stays almost entirely
under 60\,ms, while fixed Q8 and Q6 exceed it on over a hundred steps each. Choosing the schedule with our cost instead of
bytes also makes it more accurate and slightly faster on average, with a similar number of slow steps
(Appendix~\ref{app:scheduleconfirm}), and the same runtime
answers as many BFCL tool-use turns~\citep{bfcl} correctly as Q8 with lower response time
(Appendix~\ref{sec:slmruntime-full}).

\section{Related work and discussion}
\label{sec:related}
\noindent\textbf{Mixed-precision allocation.} Prior allocators choose precision from layer sensitivity under a
resource limit. HAWQ prices configurations by model size~\citep{hawq} and dMX by average bit-width~\citep{dmx}.
HAWQ-V3 showed that latency does not follow size or FLOPs and limits it with per-layer times measured on one
GPU~\citep{hawqv3}, and HAQ found with simulated latency and energy that the best policy differs by hardware and
between memory- and compute-bound layers~\citep{haq}. LLM-PQ profiles decoder layers on each GPU and fits
per-precision regressions on FLOPs in prefill and bytes in decode, after observing that lower precision can be
slower~\citep{llmpq}, and MxMoE prices formats with profiled kernel tiles~\citep{mxmoe}. Measured tables capture
kernel effects but must be rebuilt when the stack changes, while bit and byte proxies generalize but assume that
fewer bits save time. Vidur~\citep{vidur} simulates deployments from profiled operators, but without precision. Our
model is calibrated once per model and stack and then prices mixes it has not measured, MoE routing, and KV-cache
traffic across operating points, inside existing allocators.

\noindent\textbf{Runtime precision.} NestedFP~\citep{nestedfp} and DPS~\citep{dps} switch the whole model between
FP16 and FP8 when load or KV-cache pressure rises, and MorphServe~\citep{morphserve} swaps sensitive layers to 4-bit
weights under memory pressure. Nested formats~\citep{anyprecision,matquant,bitstack,apexquant} store several
precisions in one copy, which DP-LLM~\citep{dpllm} uses to pick each layer's precision per decoding step.
Disaggregated Quantization~\citep{disaggquant}, Mix-Quant~\citep{mixquant}, and PMPD~\citep{pmpd} change precision
by phase or as generation proceeds. These choose precision from accuracy, bit-width, or load thresholds, while we
choose it per layer and operating point from calibrated time costs. KIVI~\citep{kivi} and KVQuant~\citep{kvquant} compress the cache.

\noindent\textbf{Limitations.} We test the allocators derived from published methods on two models with block-scaled FP8 and one with
MXFP4. The smooth interpolation of Equation~\ref{eq:blend} can miss slowdowns when the runtime changes kernels
(Appendix~\ref{app:kernelswitch}), and one rate per format cannot express a format that is fast in one kernel and slow in
another. A per-shape latency table beats our selections for dense models under memory budgets in decode
(Appendix~\ref{app:budgets}), so making each format's calibrated bandwidth depend on matrix shape is a natural extension.
Summed per-tensor KL ranks configurations well by joint KL but overestimates the joint KL itself (Appendix~\ref{sec:opdivergence}). Another extension is to test the model on hardware from other vendors.

\section{Conclusion}
Overall, bytes are a good proxy for time in many settings, but they miss how the hardware runs each format. Across three
hardware classes, roughly a third of the measured pairs of configurations ran slower with fewer bytes. A cost model
calibrated on compute, weight loading, and KV-cache costs captures these cases, and it needs only a one-time
calibration per model, device, and serving stack.

The cost model replaces only the time estimate inside an allocator, so existing allocators can use it while keeping
their own accuracy measurements and search. With it, they select configurations with up to $52\%$ lower latency where
bytes misjudge speed, with measured accuracy within noise, and once recalibrated they follow kernel changes that
bytes cannot see. Because the model takes the operating point as input, it also shows how precision should follow
the device, batch size, phase, and context length, and switching precision as these change makes serving faster.
More broadly, how fast a precision runs is not a fixed property of its format. It depends on the hardware, the
kernels, and the operating point, and mixed-precision allocation should account for all three.

\section*{Acknowledgments}
AMD, the AMD Arrow logo, AMD Instinct, AMD ROCm, Radeon, and combinations thereof are trademarks of Advanced Micro
Devices, Inc. Other product names used in this publication are for identification purposes only and may be
trademarks of their respective companies.

\bibliographystyle{mlsys2025}
\bibliography{refs,sr_refs}

\clearpage
\appendix
\let\arxivoldsection\section
\renewcommand{\section}{\FloatBarrier\arxivoldsection}
\renewcommand{\thesection}{A\arabic{section}}
\counterwithin{figure}{section}
\counterwithin{table}{section}
\renewcommand{\thefigure}{\thesection.\arabic{figure}}
\renewcommand{\thetable}{\thesection.\arabic{table}}
\renewcommand{\theequation}{A\arabic{equation}}
We use three devices throughout, the AMD Instinct\tmark{} MI355X GPU, the AMD Instinct\tmark{} MI300A datacenter APU,
and AMD Radeon\tmark{} 780M Graphics. Sections~\ref{app:opt}--\ref{app:baselineformulas} describe the allocator,
the calibration, and the baselines. Sections~\ref{app:testing}--\ref{app:routing} test the cost model further,
Section~\ref{app:full} gives the full allocation results, and
Sections~\ref{sec:opdep-full}--\ref{app:kvcontext} give the operating-point experiments. The last sections cover the
accuracy measurements and supporting material, and Section~\ref{app:configs} collects the mixed-precision
configurations behind every main-text figure.

\section{Allocator details}
\label{app:opt}
This section shows the allocator's inputs, the accuracy--latency tradeoff, and the dynamic program it uses.

\begin{figure}[H]
    \centering
    \resizebox{0.95\textwidth}{!}{%
    \begin{tikzpicture}[
        font=\footnotesize, >={Stealth[]}, line cap=round,
        cell/.style={draw, rounded corners=0.5pt, fill=blue!4, align=center, font=\scriptsize,
                     minimum width=34mm, minimum height=4.4mm, inner sep=0.8mm},
        chip/.style={draw, rounded corners=0.5pt, align=center, font=\scriptsize,
                     minimum width=16mm, minimum height=4.4mm, inner sep=0.8mm},
        live/.style={draw, rounded corners=0.5pt, fill=orange!14, align=left, font=\scriptsize,
                     minimum width=34mm, inner sep=1mm},
        hw/.style={draw, rounded corners=0.5pt, fill=green!12, align=center, font=\scriptsize,
                   minimum width=24mm, minimum height=4.4mm, inner sep=0.8mm},
        grp/.style={draw, rounded corners, inner sep=1.3mm},
        box/.style={draw, rounded corners, fill=blue!16, align=center,
                      minimum width=24mm, minimum height=12mm, font=\small\bfseries},
        layer/.style={draw, align=center, minimum height=4.2mm, minimum width=20mm, font=\scriptsize},
        glab/.style={font=\scriptsize\bfseries},
    ]
        \colorlet{cf4}{red!20}\colorlet{cf6}{orange!35}\colorlet{cf8}{yellow!45}\colorlet{cbf}{blue!35}
        \node[cell] (c1) {compute rate $r(p)$ per format};
        \foreach \i/\p/\t in {2/c1/{bandwidth $\mathrm{BW}(p)$ per format}, 3/c2/{overheads $h(u,p)$},
                              4/c3/{blend scale $M_{1/2}$}, 5/c4/{baseline $f_\text{base}$},
                              6/c5/{attention $a_\text{prefill}, a_\text{decode}$}}
            \node[cell, below=0.7mm of \p] (c\i) {\t};
        \node[grp, fit=(c1)(c6), label={[glab]above:Calibrated per model, device, and stack}] (cal) {};
        \node[live, below=9mm of c6] (op) {hardware and serving stack\\phase (prefill or decode)\\batch size\\prompt length\\context length};
        \node[grp, fit=(op), label={[glab]above:Operating point}] (opg) {};
        \node[box, right=12mm of cal.east, yshift=-10mm] (cm) {Cost model\\[0.5mm]{\normalfont\scriptsize time change $\Delta S(u,p)$}};
        \node[cell, above=12mm of cm, xshift=26mm] (acc) {accuracy table $\Delta A(u,p)$};
        \node[grp, fit=(acc), label={[glab]above:Measured once per model}] (accg) {};
        \node[cell, right=6mm of acc, minimum width=30mm] (bud) {memory, speed, or accuracy};
        \node[grp, fit=(bud), label={[glab]above:Budget}] (budg) {};
        \node[box, right=14mm of cm] (al) {Allocator\\[0.5mm]{\normalfont\scriptsize ours or published}};
        \node[chip, fill=cbf, below=24mm of al, xshift=-16mm] (s1) {BF16};
        \foreach \i/\p/\c/\t in {2/s1/cf8/FP8, 3/s2/cf8/MXFP8, 4/s3/cf6/MXFP6, 5/s4/cf4/MXFP4}
            \node[chip, fill=\c, right=0.7mm of \p] (s\i) {\t};
        \node[grp, fit=(s1)(s5), label={[glab]below:Formats the stack supports (example)}] (menu) {};
        \node[layer, fill=cbf, right=14mm of al, yshift=8mm] (m1) {BF16};
        \foreach \i/\p/\c/\t in {2/m1/cf6/MXFP6, 3/m2/cf4/MXFP4, 4/m3/cf8/FP8}
            \node[layer, fill=\c, below=1mm of \p] (m\i) {\t};
        \node[layer, minimum height=3mm, below=1mm of m4] (md) {$\vdots$};
        \node[grp, fit=(m1)(md), label={[glab]above:Per-layer configuration}] (model) {};
        \node[hw, right=8mm of model, minimum width=22mm, minimum height=8mm] (g1) {served on\\the device};
        \draw[->] (cal.east) -- (cm.west |- cal.east) -- (cm.north west);
        \draw[->] (opg.east) -| (cm.south);
        \draw[->, thick] (cm.east) -- (al.west);
        \draw[->] (accg.south) -- (al.north west);
        \draw[->] (budg.south) -- (al.north east);
        \draw[->] (menu.north -| al.south) -- (al.south);
        \draw[->, thick] (al.east) -- node[above, font=\scriptsize]{config} (model.west);
        \draw[->, thick] (model.east) -- (g1.west);
        \draw[<-, dashed, orange!80!black] (opg.south) -- ++(0,-17mm) -| node[pos=0.25, below, font=\scriptsize]{when the operating point changes, re-run the cost model and allocator} (g1.south);
    \end{tikzpicture}}
    \caption{Method overview. The calibrated cost model gives a time change for each unit and format, and the
    allocator uses it with the accuracy table and budget to pick a precision for each layer.}
    \label{fig:arch}
\end{figure}

Figure~\ref{fig:arch} shows the inputs to the allocator, and Figure~\ref{fig:optline} illustrates the tradeoff
it makes for each unit.
Algorithm~\ref{alg:allocate} first adds up each unit's minimum time change $m_u$ to get $m$, the time of the fastest
modeled configuration. Each time bin is an amount of extra time above it. For each unit, the algorithm tries every
precision and keeps the lowest accuracy cost in each bin. It then picks the best entry within the latency or accuracy
budget and backtracks to recover the choices. We use 20{,}000 bins on MI355X. Since times are rounded up, some
feasible choices can be lost, so we check the selected configuration against the original budget.

\begin{algorithm}[H]
\caption{Precision allocation by dynamic programming}
\label{alg:allocate}
\small
\begin{algorithmic}[1]
\REQUIRE Units $u=1,\ldots,n$, precision menus $\mathcal{P}_u$, accuracy changes $\Delta A(u,p)$,
time changes $\Delta S(u,p)$, grid width $\delta>0$, and latency budget $\tau$ or accuracy budget $\epsilon$
\STATE $m_u\gets\min_{p\in\mathcal{P}_u}\Delta S(u,p)$; $m\gets\sum_u m_u$
\STATE $c(u,p)\gets\left\lceil(\Delta S(u,p)-m_u)/\delta\right\rceil$ for every option
\STATE $C\gets\sum_u\max_{p\in\mathcal{P}_u}c(u,p)$
\IF{latency budget}
  \STATE $C\gets\min\!\left(C,\left\lfloor(\tau-m)/\delta\right\rfloor\right)$
  \IF{$C<0$} \STATE \textbf{return}\ infeasible \ENDIF
\ENDIF
\STATE $D[0]\gets0$ and $D[j]\gets+\infty$ for $j=1,\ldots,C$
\FOR{$u=1,\ldots,n$}
  \FOR{$j=0,\ldots,C$}
    \STATE $D_{\mathrm{new}}[j]\gets\min_{p\in\mathcal{P}_u:\,c(u,p)\le j}
      \bigl(D[j-c(u,p)]+\Delta A(u,p)\bigr)$
    \STATE Record a minimizing precision and preceding bin whenever the value is finite
  \ENDFOR
  \STATE $D\gets D_{\mathrm{new}}$
\ENDFOR
\STATE $\mathcal{F}\gets\{j:D[j]<+\infty\}$
\IF{accuracy budget} \STATE $\mathcal{F}\gets\{j\in\mathcal{F}:D[j]\le\epsilon\}$ \ENDIF
\IF{$\mathcal{F}=\varnothing$} \STATE \textbf{return}\ infeasible \ENDIF
\STATE Choose $j^\star\in\arg\min_{j\in\mathcal{F}}D[j]$ for a latency budget, or $j^\star=\min\mathcal{F}$ for an accuracy budget
\STATE Backtrack the recorded choices from $j^\star$ to recover one precision $p_u$ per unit
\STATE Verify the original unrounded constraint and \textbf{return}\ $\{p_u\}$
\end{algorithmic}
\end{algorithm}
Algorithm~\ref{alg:allocate} discretizes time. The allocations in Section~\ref{sec:existing} discretize the
accuracy cost instead and track the lowest time. Both are exact for the discretized additive problem.

\begin{figure}[H]
\centering
\resizebox{0.5\linewidth}{!}{%
\begin{tikzpicture}[font=\footnotesize, >={Stealth[round]}, line cap=round]
  \fill[green!8] (0,0) -- (0,5.2) -- (1.0,5.2) -- (6.2,0) -- cycle;
  \draw[->, thick] (0,0) -- (7.2,0);
  \node[below] at (5.6,-0.05) {accuracy impact $\Delta A(u)$};
  \draw[->, thick] (0,0) -- (0,5.5);
  \node[above] at (0,5.5) {time change $\Delta S(u)$};
  \draw[dashed, gray!75] (1.0,5.2) -- (6.2,0);
    \node[gray!50!black, rotate=-45, font=\scriptsize] at (4.55,1.2) {equal cost: $\Delta A{+}\lambda\Delta S{=}c$};
  \fill[blue!55!black] (3.6,2.6) circle (2.4pt);
  \node[anchor=north west, font=\scriptsize] at (3.7,2.55) {unit $u$: $(\Delta A,\Delta S)$};
  \node[green!45!black, align=center] at (1.35,0.85) {beneficial\\region};
  \draw[->, green!50!black, very thick] (3.6,2.6) -- (2.1,3.0);
  \node[green!45!black, anchor=east, font=\scriptsize] at (2.05,3.05) {lower cost};
  \draw[->, red, very thick] (3.6,2.6) -- (3.25,3.95);
  \node[red, anchor=south, font=\scriptsize] at (3.25,4.0) {higher cost};
\end{tikzpicture}}
\caption{Quantizing a unit moves it in accuracy impact and time change. Moves toward the shaded region give a
better tradeoff, and the best move differs between units, so each unit gets its own precision.}
\label{fig:optline}
\end{figure}

\section{Calibrating the cost model}
\label{app:roofline}
The cost model of Section~\ref{sec:method} has a small set of parameters fitted once per deployment. This section
explains its roofline form, how we fit it, and the fitted values.

Our model uses the roofline approximation to separate the cost of arithmetic from the cost of loading weights
\citep{scalingbook,llmviewer}.

\paragraph{Expected active experts.} A token skips a given expert with probability $1-k/E$, so under independent
uniform routing the expert receives at least one of the $BS$ tokens with probability $1-(1-k/E)^{BS}$. Summing over
experts gives Equation~\ref{eq:activeexperts}, and dividing the $BSk$ assignments by $E_+$ gives the balanced-load
approximation.

\paragraph{Why use a maximum?} A tiled matrix multiplication can load the next tile while it computes the current
one. With ideal overlap, its steady-state time is about
\begin{equation}
T\approx\max\big(\underbrace{2WM/r}_{\text{compute}},\ \underbrace{Wb/\mathrm{BW}}_{\text{weight load}}\big),
\label{eq:roofline1}
\end{equation}
where $W$ is the number of weights, $M$ the number of token rows, $b$ the bytes per weight, $r$ the throughput,
and $\mathrm{BW}$ the bandwidth. More rows mean more reuse of each weight, so the limit can move from memory to
compute. Overheads and other operations get their own terms.

\paragraph{Calibration procedure.}
All platforms use the same cost equation and fitter. We average repeated timings for each configuration and
workload. On the datacenter devices we measure whole requests, so prefill time is the latency of a request with one output
token, and decode time per step is how much latency grows with each additional output token. On the laptop,
llama.cpp reports both phases directly.
For prefill and decode calibration sets $\mathcal C_{\mathrm p}$ and $\mathcal C_{\mathrm d}$, we minimize the
equally weighted mean squared relative errors

\begin{equation}
\begin{aligned}
J_{\mathrm{base}}&=\tfrac12 P+\tfrac12 D,\\
P&=\frac{1}{|\mathcal C_{\mathrm p}|}\sum_{i\in\mathcal C_{\mathrm p}}
\left(\frac{\widehat T_i-T_i}{T_i}\right)^2,\\
D&=\frac{1}{|\mathcal C_{\mathrm d}|}\sum_{i\in\mathcal C_{\mathrm d}}
\left(\frac{\widehat T_i-T_i}{T_i}\right)^2.
\end{aligned}
\label{eq:calibrationloss}
\end{equation}
Here $P$ and $D$ are the mean squared relative errors for prefill and decode. We weight them equally so that
neither phase counts more just because it has more measurements.

We fit one parameter set per deployment. Rates, bandwidths, $M_{1/2}$, and overheads are shared by prefill and
decode, so both phases constrain them, while baseline and attention costs are fitted per phase. The fitted values are
effective rather than physical, because a term that rarely limits time is barely constrained by the data and can
exceed the device's peak (Table~\ref{tab:effbw_params}). Table~\ref{tab:effbw_shared} gives the fitted baseline
and attention costs.

\begin{table}[H]
\centering\small
\setlength{\tabcolsep}{3pt}
\caption{Fitted parameters of our model (one effective bandwidth per format), calibration cells only. Rates in TFLOP/s, bandwidths in TB/s; shared BW = the single bandwidth fitted for all formats in the ablation; $M_{1/2}$ = blend scale (rows); MAPE on calibration cells (prefill / decode); converged = how many of 32 fits from random starting points converged. $\dagger$: at a bound (not identified); $\ddagger$: column inactive or collinear (cosine $>0.95$) with a per-format overhead.}
\label{tab:effbw_params}
\begin{tabular}{@{}llrrrrrc@{}}\toprule
Stack & Format & Rate & Per-format BW & Shared BW & $M_{1/2}$ & MAPE (\%) & Converged\\\midrule
MI355X, Qwen3-4B (v0.25) & BF16 & 1,202 & 2.5 & 2.5 & 138 & 9.7 / 7.8 & 17\\
 & MXFP4 & 4,405 & 100.0$^\dagger$ &  &  &  & \\
\addlinespace[2pt]
MI355X, Coder-30B (v0.25) & BF16 & 869 & 9.1 & 9.4 & 2,468 & 10.5 / 3.1 & 32\\
 & MXFP4 & 10,000$^\dagger$ & 6.3 &  &  &  & \\
\addlinespace[2pt]
MI355X, Coder-30B nat. & BF16 & 882 & 9.2 & 9.5 & 2,596 & 10.5 / 3.1 & 32\\
 & MXFP4 & 10,000$^\dagger$ & 6.2 &  &  &  & \\
\addlinespace[2pt]
MI355X, Coder-30B bal. & BF16 & 726 & 5.3 & 5.2 & 3,244 & 11.2 / 2.3 & 25\\
 & MXFP4 & 10,000$^\dagger$ & 4.3 &  &  &  & \\
\addlinespace[2pt]
MI300A, Qwen2.5-7B & BF16 & 198 & 100.0$^\dagger$ & 100.0$^\dagger$ & 545 & 4.8 / 15.7 & 32\\
 & FP8 & 739 & 2.0 &  &  &  & \\
\addlinespace[2pt]
780M, Llama-3.2-1B & F16 & 0.8 & 0.28 & 0.17 & 0.001$^\dagger$ & 6.6 / 3.9 & 31\\
 & Q8\_0 & 1.3 & 0.75 &  &  &  & \\
\addlinespace[2pt]
780M, five formats & F16 & 0.57 & 0.048 & 0.06 & 0.001$^\dagger$ & 5.2 / 3.2 & 32\\
 & Q4\_K & 0.97 & 0.042 &  &  &  & \\
 & Q5\_K & 0.81 & 0.18 &  &  &  & \\
 & Q6\_K & 0.62 & 0.31 &  &  &  & \\
 & Q8\_0 & 1.1 & 0.03 &  &  &  & \\
\addlinespace[2pt]
MI355X, Coder-30B (v0.31) & BF16 & 503 & 10.1 & 10.1 & 531 & 14.7 / 3.0 & 32\\
 & MXFP4 & 1,663 & 8.4 &  &  &  & \\
\addlinespace[2pt]
MI355X, Qwen3-4B (v0.31) & BF16 & 870 & 2.9 & 2.9 & 0.001$^\dagger$ & 9.3 / 8.2 & 26\\
 & MXFP4 & 2,382 & 0.91$^\ddagger$ &  &  &  & \\
\addlinespace[2pt]
\bottomrule
\end{tabular}
\end{table}

\begin{table}[H]
\centering\small
\setlength{\tabcolsep}{4pt}
\caption{Fitted baseline and attention costs of our model. The baselines cover work that does not depend on precision, such as normalization and serving-engine overhead. Prefill attention is per million query--key pairs and decode attention per 1{,}024 attended tokens. $^c$: prefill baseline charged per prefill chunk.}
\label{tab:effbw_shared}
\begin{tabular}{@{}lrrrr@{}}\toprule
& \multicolumn{2}{c}{Baseline $f_\text{base}$ (ms)} & \multicolumn{2}{c}{Attention (ms)}\\\cmidrule(lr){2-3}\cmidrule(l){4-5}
Stack & Prefill & Decode & Prefill & Decode\\\midrule
MI355X, Qwen3-4B (v0.25) & 20.1 & 0 & 0.499 & 0.106\\
MI355X, Coder-30B (v0.25) & 39.3 & 3.8 & 0.260 & 0.011\\
MI355X, Coder-30B nat. & 41.9 & 3.9 & 0.249 & 0.011\\
MI355X, Coder-30B bal. & 38.1 & 3.4 & 0.133 & 0.017\\
MI300A, Qwen2.5-7B & 14.0$^c$ & 10.9 & 1.6 & 0.039\\
780M, Llama-3.2-1B & 37.3$^c$ & 35.2 & 239.5 & 2.6\\
780M, five formats & 53.5$^c$ & 0 & 295.4 & 3.2\\
MI355X, Coder-30B (v0.31) & 35.8 & 3.4 & 0 & 0\\
MI355X, Qwen3-4B (v0.31) & 16.7 & 0 & 0 & 0.063\\
\bottomrule
\end{tabular}
\end{table}

\paragraph{Generating evaluation configurations.}
Each dense-model layer has three groups of projections, Q/K/V, attention output, and feed-forward. We generate six
configurations by quantizing each group independently with probability 10\%, 25\%, 40\%, 60\%, 75\%, and 90\%,
and freeze them before measuring. For the MoE model, we test 28 held-out assignments that span cost, storage, and
sensitivity budgets, and BF16 plus three random mixes across workloads.

\section{Byte and constant latency estimates}
\label{app:baselineformulas}
We compare our model with two simpler estimates fitted on the same calibration data.

Both baselines predict latency as a weighted sum of known inputs with fitted coefficients,
\[
\widehat T_{\mathrm{constant}}=\boldsymbol\theta_{\mathrm c}^{\mathsf T}\boldsymbol x_{\mathrm c},
\qquad
\widehat T_{\mathrm{byte}}=\boldsymbol\theta_{\mathrm b}^{\mathsf T}\boldsymbol x_{\mathrm b}.
\]
Each baseline fits its own coefficients to the same calibration timings. The constant estimate changes with the
workload but not with precision. Here $B$ is the batch size, $S$ the prompt length, $C$ the number of executed prefill
chunks, and $\overline L$ the effective decode context. Only $G$ depends on precision. It is the size of the
allocatable weights in GB, including block metadata and all stored MoE experts. $I_{\mathrm p}$ is 1 for prefill
and $I_{\mathrm d}=1-I_{\mathrm p}$.

\paragraph{Shared input terms.}
All platforms and model types use
\[
\begin{aligned}
\boldsymbol x_{\mathrm c}
 &=\Bigl(I_{\mathrm p},\ I_{\mathrm p}C,\ I_{\mathrm p}B,\
          \frac{I_{\mathrm p}BS}{1024},\\
 &\qquad\frac{I_{\mathrm p}BS^2}{10^6},\ I_{\mathrm d},\ I_{\mathrm d}B,\
          \frac{I_{\mathrm d}B\overline L}{1024}\Bigr)^{\mathsf T},\\[3pt]
\boldsymbol x_{\mathrm b}
 &=\left(\boldsymbol x_{\mathrm c}^{\mathsf T},\ I_{\mathrm p}GC,\ I_{\mathrm d}G,\
          \frac{I_{\mathrm p}GBS}{1024}\right)^{\mathsf T}.
\end{aligned}
\]
Datacenter prefill uses $C=1$, the laptop uses $B=1$ and $C=\lceil S/512\rceil$, and decode contexts match our
model. We fit nonnegative coefficients that minimize Equation~\ref{eq:calibrationloss}, so fewer stored bytes can
never increase the byte estimate at a fixed workload.

\paragraph{MI355X MoE decode example.}
In Figure~\ref{fig:bytetracking}, our model expects $128[1-(1-8/128)^{16}]=82.42$ active experts per layer. Expert
weight traffic grows with batch size, but the byte estimate's decode precision term is just $\beta_GG$, so it gives
the same saving at every batch size. From BF16 to configuration~1, measured latency falls by $2.08$\,ms, our model
predicts $1.71$\,ms, and the byte estimate predicts $0.95$\,ms. In the fitted model all operators are memory-bound
here, so the difference comes from expert traffic.

For dense MI355X decode, bytes barely affect latency, so the byte estimate fits a near-zero storage coefficient and
behaves like the constant estimate. Our model follows the measured latency across workloads more closely
(Table~\ref{tab:proxies}).

\paragraph{Context scheduling.}
For batch-one decode scheduling, the byte estimate is
\[
\begin{gathered}
\widehat T_{\mathrm{byte,schedule}}=\beta_0+\beta_L\overline L/1024+\beta_GG,\\
\beta_0,\beta_L,\beta_G\ge0.
\end{gathered}
\]
We fit both this estimate and our model with the paired-difference objective of Appendix~\ref{sec:hardware}.

\section{Testing the cost model further}
\label{app:testing}
This section tests the cost model beyond the average error in Section~\ref{sec:predict}. We check how often it
catches fewer bytes running slower, its accuracy on every workload and configuration, and how much its selections
depend on the calibration set.

\subsection{Fewer bytes running slower}

\begin{figure}[H]
\centering
\includegraphics[width=0.95\linewidth,height=0.45\textheight,keepaspectratio]{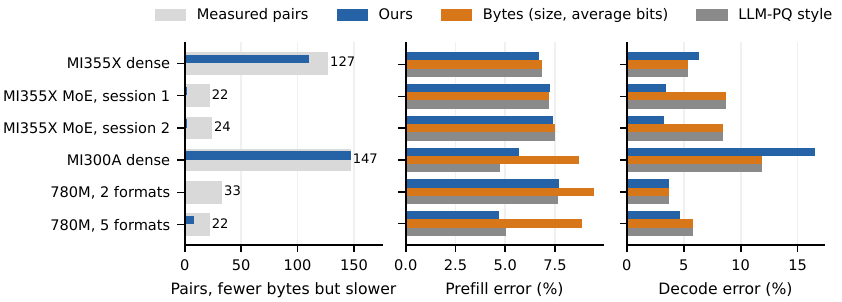}
\caption{Measured pairs where the configuration with fewer bytes ran more than $2\%$ slower and the number each
estimate predicts (left), and held-out prediction error (middle and right).}
\label{fig:c1}
\end{figure}

The LLM-PQ-style estimate follows the per-precision latency regressions of LLM-PQ~\citep{llmpq}. Fitted on our
calibration data, it predicts time from FLOPs in prefill and bytes in decode. Figure~\ref{fig:c1} and
Table~\ref{tab:proxies} compare our prediction error with the byte and LLM-PQ-style estimates. An inversion is a measured pair at the same workload where the
configuration with fewer stored bytes ran more than $2\%$ slower. Byte estimates predict none of them and
LLM-PQ-style estimates predict 5. Our model predicts 269 of 375, with precision 0.64. Counting only pairs whose sign
agrees in every block (931 pairs, 316 inversions), precision is 0.76 and recall 0.81, against 0.68 and 0.48 with one
shared bandwidth. Recall is high where the format itself is slow (MI300A and MI355X dense) and near zero on MoE and
the Radeon~780M.

The result holds when we pool every configuration we served, including the allocation experiments and fresh
held-out configurations. Our model then catches 519 of 649 resolved pairs where fewer bytes ran slower (recall 0.80,
precision 0.78), against recall 0.54 with one shared bandwidth.
\begin{table}[H]
\centering\small
\caption{Held-out latency prediction of our model, our model with one shared bandwidth, bytes, and an LLM-PQ-style
regression. Inversions are pairs where fewer stored bytes ran more than $2\%$ slower.}
\label{tab:proxies}
\setlength{\tabcolsep}{3.5pt}
\begin{tabular}{llccccc}\toprule
 & & \multicolumn{2}{c}{MAPE (\%) / order accuracy} & & \multicolumn{2}{c}{Inversions}\\\cmidrule(lr){3-4}\cmidrule(lr){6-7}
Platform, model & Phase & Ours / shared & Bytes / LLM-PQ & & Count (max) & Ours / shared\\\midrule
MI355X, Qwen3-4B & prefill & 6.7/0.78 \;\; 6.6/0.77 & 6.8/0.73 \;\; 6.8/0.73 & & 127 (18.6\%) & 110 / 102\\
 & decode & 6.3/0.67 \;\; 6.4/0.67 & 5.4/0.67 \;\; 5.4/0.67 & & & \\
MI355X, Qwen3-Coder-30B-A3B & prefill & 7.3/0.29 \;\; 7.2/0.29 & 7.2/0.24 \;\; 7.2/0.24 & & 46 (27.0\%) & 4 / 4\\
 & decode & 3.5/1.00 \;\; 3.3/1.00 & 8.7/1.00 \;\; 8.7/1.00 & & & \\
\quad balanced routing & prefill & 8.0/0.25 \;\; 7.9/0.25 & 7.6/0.19 \;\; 7.6/0.19 & & \multicolumn{2}{c}{not counted}\\
 & decode & 1.1/1.00 \;\; 1.3/1.00 & 16.1/1.00 \;\; 16.1/1.00 & & & \\
MI300A, Qwen2.5-7B & prefill & 5.7/1.00 \;\; 5.9/1.00 & 8.7/1.00 \;\; 4.7/1.00 & & 147 (40.4\%) & 147 / 49\\
 & decode & 16.5/0.96 \;\; 13.9/0.36 & 11.9/0.96 \;\; 11.9/0.96 & & & \\
Radeon 780M, Llama-3.2-1B & prefill & 7.7/0.95 \;\; 7.7/0.95 & 9.4/0.95 \;\; 7.7/0.95 & & 33 (14.8\%) & 0 / 0\\
 & decode & 3.7/0.79 \;\; 3.7/0.79 & 3.7/0.79 \;\; 3.7/0.79 & & & \\
\quad five formats & prefill & 4.7/0.49 \;\; 4.6/0.49 & 8.8/0.57 \;\; 5.0/0.37 & & 22 (17.3\%) & 8 / 8\\
 & decode & 4.7/0.83 \;\; 4.5/0.83 & 5.8/0.81 \;\; 5.8/0.81 & & & \\\bottomrule
\end{tabular}
\end{table}

\subsection{Every workload and configuration}

\begin{figure}[H]
\centering
\includegraphics[width=0.95\linewidth,height=0.78\textheight,keepaspectratio]{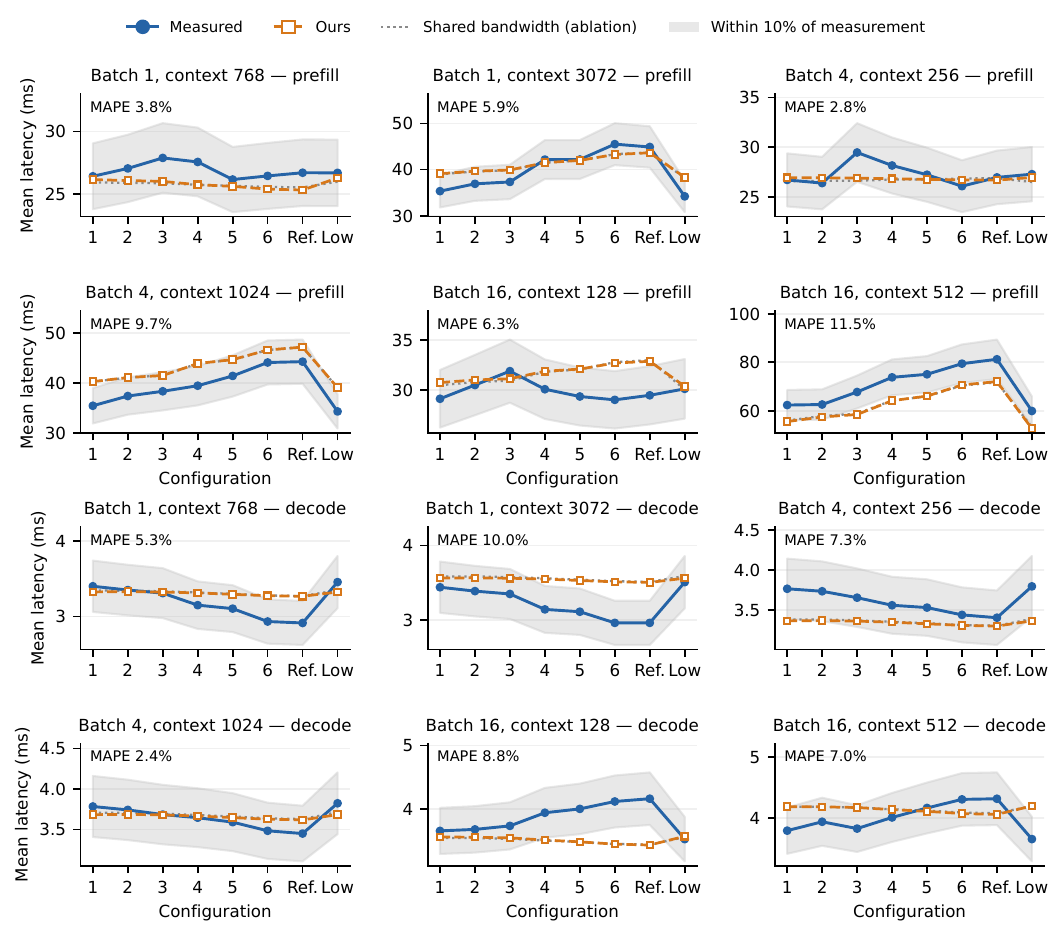}
\caption{Measured and predicted latency of the dense model on MI355X (eight configurations, six held-out workloads).
Gray bands mark $\pm10\%$ of measured latency.}
\label{fig:densedetail}
\end{figure}

\begin{figure}[H]
\centering
\includegraphics[width=0.95\linewidth,height=0.78\textheight,keepaspectratio]{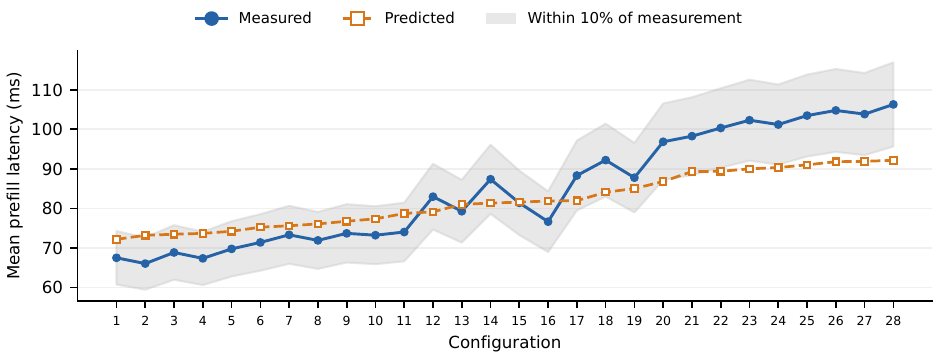}
\caption{Measured and predicted prefill latency of 28 held-out MoE configurations on MI355X, each averaged over batch~32 with 256-token prompts and batch~8 with 1024-token prompts.}
\label{fig:moetracking}
\end{figure}

\begin{figure}[H]
\centering
\includegraphics[width=0.95\linewidth,height=0.78\textheight,keepaspectratio]{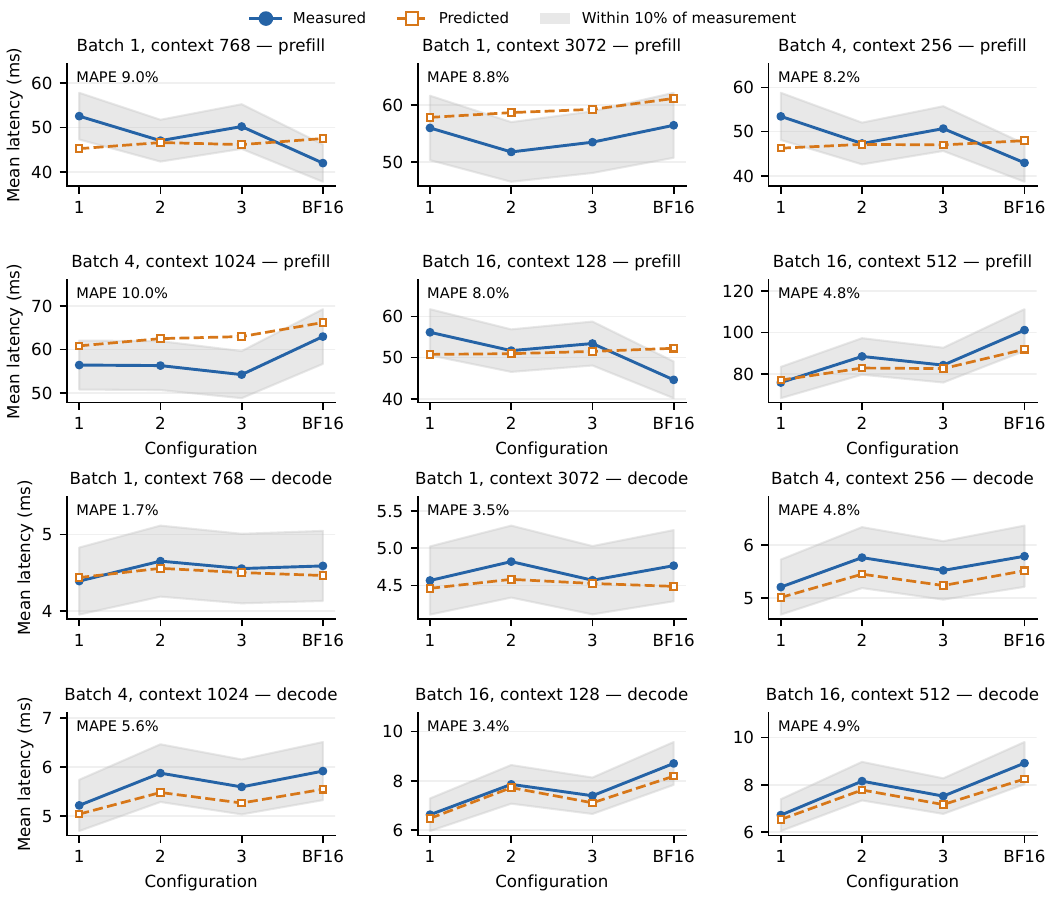}
\caption{Measured and predicted latency of the MoE model on MI355X (four configurations, six held-out workloads).}
\label{fig:predmi355detail}
\end{figure}

\begin{figure}[H]
\centering
\includegraphics[width=0.95\linewidth,height=0.78\textheight,keepaspectratio]{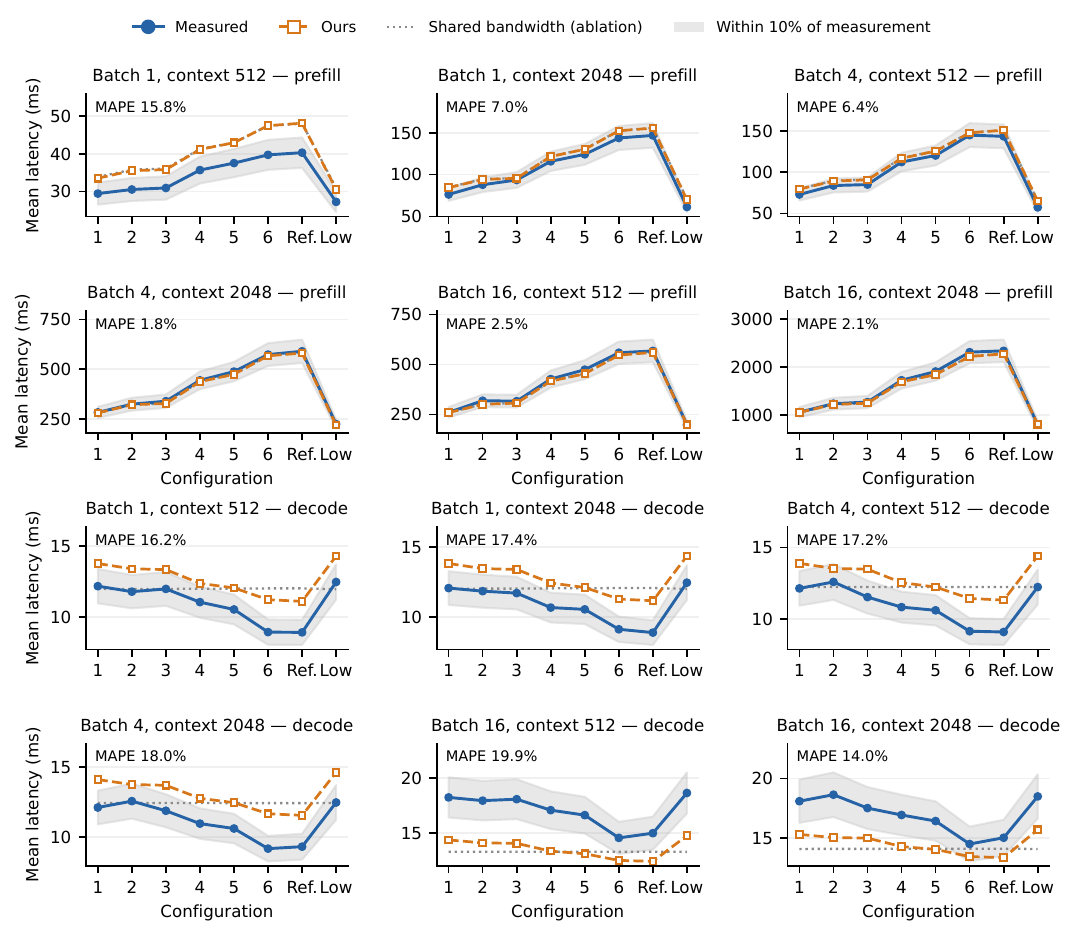}
\caption{Measured and predicted latency on MI300A (eight configurations, six workloads). Gray bands mark $\pm10\%$.}
\label{fig:predmi300detail}
\end{figure}

\begin{figure}[H]
\centering
\includegraphics[width=0.95\linewidth,height=0.78\textheight,keepaspectratio]{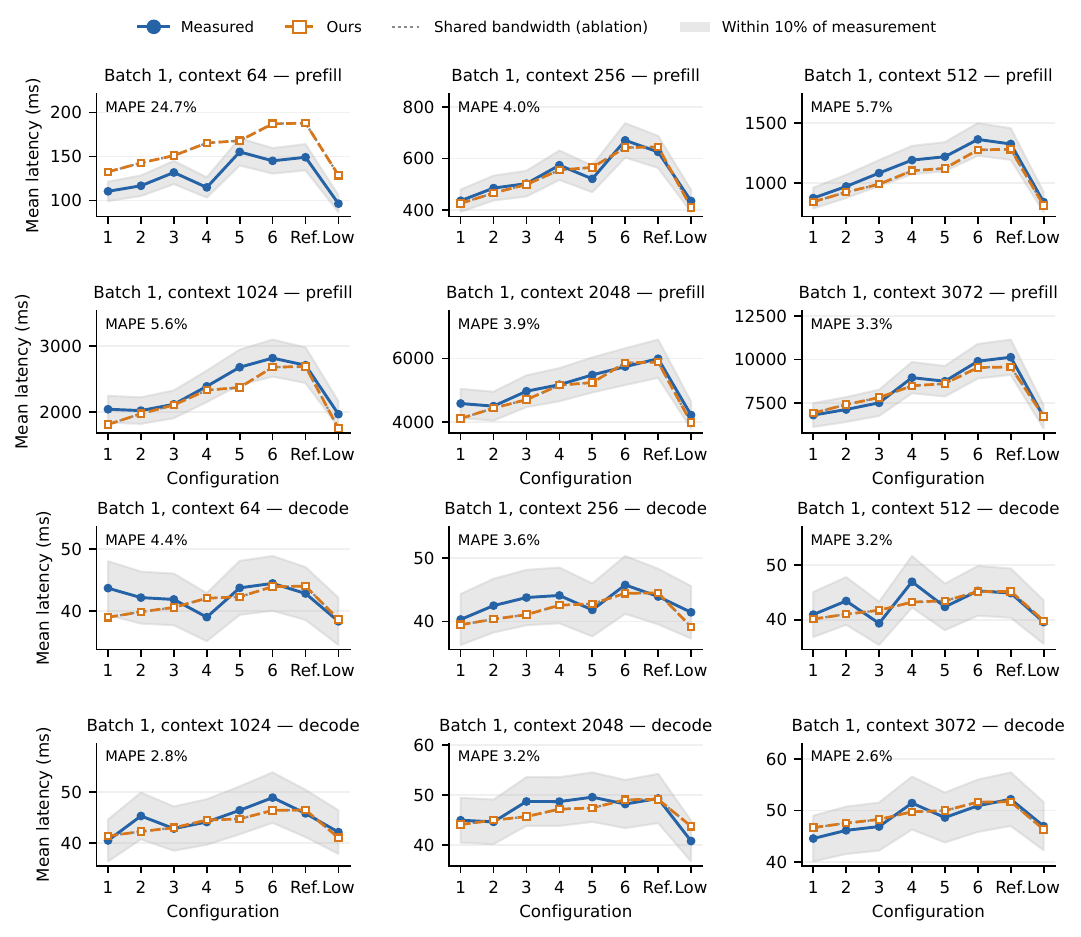}
\caption{Measured and predicted latency on the Radeon~780M (eight configurations, six contexts).}
\label{fig:predlaptopdetail}
\end{figure}

Figures~\ref{fig:densedetail}--\ref{fig:predlaptopdetail} and
Table~\ref{tab:mi355validation} test the cost model on every model, device, and operating point we evaluate. In
these figures, configurations 1--6 are those of Figure~\ref{fig:predmeas}, and the controls Ref. and Low keep all
projections at reference or quantized precision and are measured at the held-out workloads.


\begin{table}[H]\centering\small
\caption{Mean absolute percentage error on MI355X at six held-out workloads, including the two control
configurations.}
\label{tab:mi355validation}
\begin{tabular}{llrr}\toprule Model & Predictor & Prefill & Decode \\ \midrule
Dense & Our model & 6.68\% & 6.80\% \\
 & Byte estimate & 6.96\% & 5.96\% \\
 & Constant estimate & 7.67\% & 5.96\% \\\midrule
MoE & Our model & 8.14\% & 3.99\% \\
 & Byte estimate & 8.30\% & 9.69\% \\
 & Constant estimate & 8.24\% & 11.04\% \\
\bottomrule\end{tabular}
\end{table}
Low average error does not always mean the model picks the faster of two configurations. For the four MoE
configurations in Figure~\ref{fig:predmi355detail}, it does so in only 39\% of prefill pairs, but in the
28-configuration test in 94--95\%. We therefore measure the selected allocations directly instead of relying on
average error.
On MI300A, our model orders 96\% of resolved decode pairs correctly but has higher decode error than the byte
estimate (Table~\ref{tab:proxies}), since it overpredicts decode at batches 1 and 4 and underpredicts
it at batch~16.

\subsection{Smaller calibration sets}
\label{app:calrobust}

Refitting the Qwen3-4B block-scaled FP8 models with one workload left out changes 0 of 288 accuracy-budget
selections on MI355X and 3 of 288 on MI300A. Fitting on half of the calibration configurations leaves most
selections unchanged, and most changes come from dropping the uniform FP8 configuration, so calibration should
include each uniform low-precision configuration.

\section{Limitation: abrupt kernel switches}
\label{app:kernelswitch}
Equation~\ref{eq:blend} moves smoothly between the reference and format rates as token rows grow. Real backends can
instead switch kernels at a fixed size, which the model does not represent well.

\begin{figure}[H]
\centering
\includegraphics[width=0.95\linewidth,height=0.78\textheight,keepaspectratio]{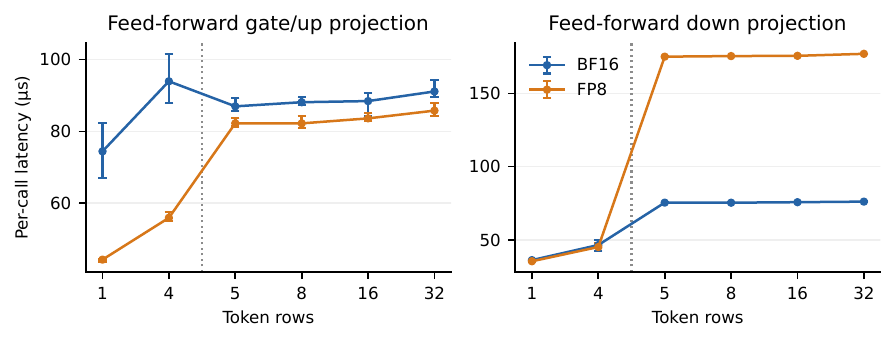}
\caption{Isolated kernel latency of two Qwen2.5-7B feed-forward projections on MI300A, averaged over six timing
groups (bars show their range). The dotted line marks the kernel switch above four token rows.}
\label{fig:mi300kernel}
\end{figure}

On MI300A, the backend switches from specialized to general matrix-multiplication kernels when a projection
processes more than four token rows, and the FP8 down projection becomes abruptly slower
(Figure~\ref{fig:mi300kernel}). This jump explains much of the batch-16 decode error.

\section{Evaluating the balanced-routing approximation}
\label{app:routing}
For MoE models, the cost model assumes each token picks experts uniformly at random, so work spreads evenly over the
active experts. Real routing is uneven, so we test how much this assumption costs.

\begin{figure}[H]
\centering
\includegraphics[width=0.95\linewidth,height=0.78\textheight,keepaspectratio]{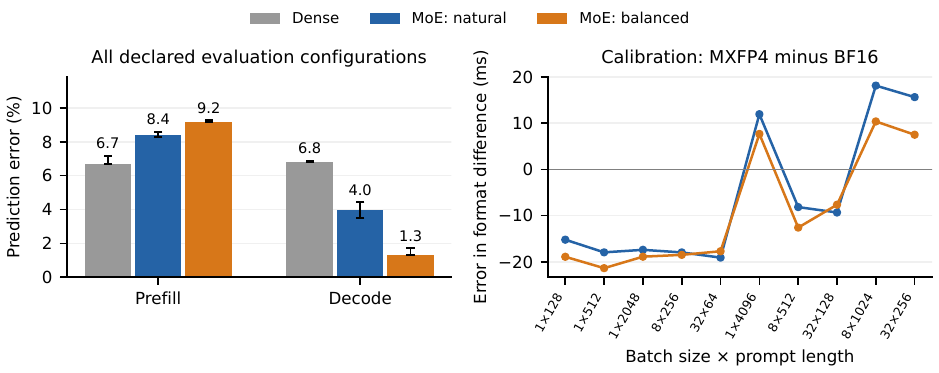}
\caption{Prediction error with normal and forced balanced MoE routing on MI355X. Left: error at six held-out
workloads, with whiskers spanning the pooled result and both model loads. Dense-model error is shown for reference.
Right: error in the predicted MXFP4--BF16 prefill gap during calibration.}
\label{fig:controlledrouting}
\end{figure}

We test the approximation in two ways. First, we force tokens to spread evenly across experts, so that routing
matches our approximation, and calibrate the model separately for normal and forced routing
(Figure~\ref{fig:controlledrouting}). Forced routing lowers decode error from 4.0\% to 1.3\% but not prefill error,
which goes from 8.4\% to 9.2\%.

\begin{figure}[H]
\centering
\includegraphics[width=0.95\linewidth,height=0.78\textheight,keepaspectratio]{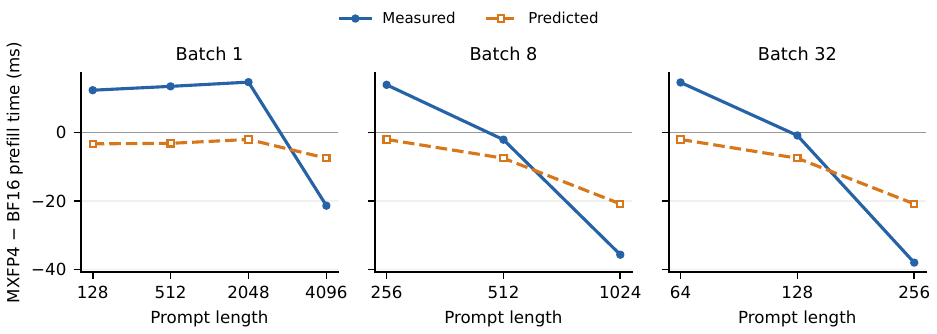}
\caption{Prefill latency of MXFP4 minus BF16 for the MoE model on MI355X (calibration workloads). Negative values mean
MXFP4 is faster. Each point averages two model loads with six requests each.}
\label{fig:formatgap}
\end{figure}

Most of the prefill error comes from understating the MXFP4--BF16 gap. MXFP4 is slower than BF16 at small
workloads, where each expert sees few tokens, and faster at large ones, and our predicted gap follows this trend but
is too flat (Figure~\ref{fig:formatgap}).

\begin{figure}[H]
\centering
\includegraphics[width=0.95\linewidth,height=0.78\textheight,keepaspectratio]{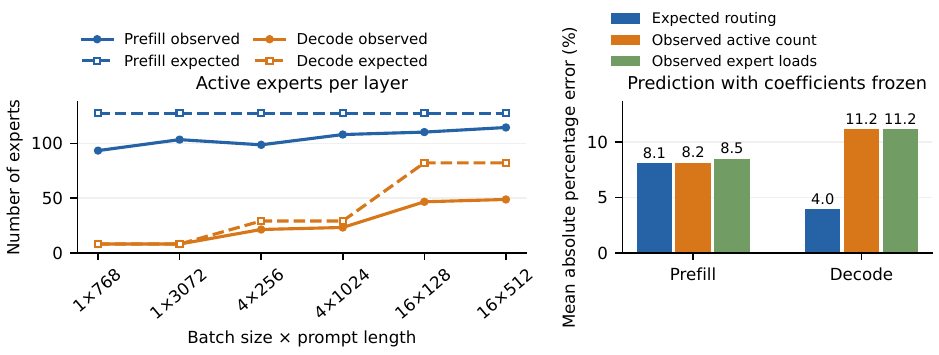}
\caption{Estimated and measured MoE routing on MI355X. Left: active experts per layer under uniform routing (expected) and as measured. Right: prediction error with fitted parameters unchanged, given the uniform-routing estimate (expected
routing), the measured number of active experts (observed active count), or the measured tokens per expert
(observed expert loads). Routes are recorded in separate runs from the timing runs.}
\label{fig:routingablation}
\end{figure}

Second, we check the active-experts estimate of Equation~\ref{eq:activeexperts} directly
(Figure~\ref{fig:routingablation}). It matches the measured count in prefill, but in decode at larger batches tokens
share experts more than uniform routing assumes, so the estimate is too high. Giving the model the measured routing
with its fitted parameters unchanged does not reduce its error. Prefill error stays near 8\%, and decode error rises
from 4.0\% to 11.2\%, since the parameters were fitted with the estimate. The remaining MoE error therefore does not
come from routing, and the uniform approximation is enough for prediction.

\section{Swapping the time cost: full results}
\label{app:full}
This section gives the full results of Section~\ref{sec:existing} for our own allocator, the allocators derived
from HAWQ-V3 and LLM-PQ, and a kernel change.

\subsection{Our allocator}
\label{app:ownalloc}
\begin{figure}[H]
\centering
\includegraphics[width=0.95\linewidth,height=0.45\textheight,keepaspectratio]{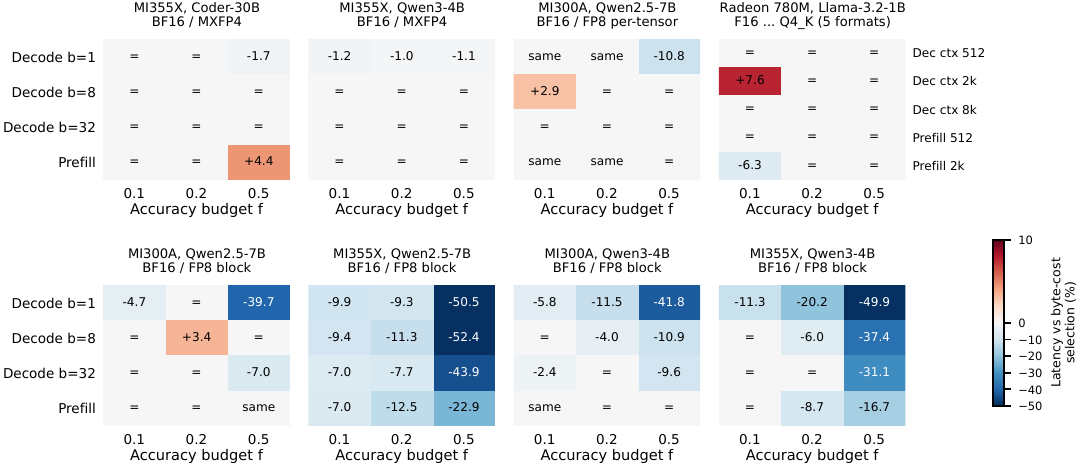}
\caption{Latency of our selection relative to the byte-cost selection at each operating point and budget $f$ (blue
is faster, $=$ means no difference outside the paired 95\% interval, Same means identical selections).}
\label{fig:c2-full}
\end{figure}

Section~\ref{sec:accbudget} summarizes what happens when an allocator uses our cost instead of bytes, and
Figure~\ref{fig:c2-full} gives every model, device, and budget. The budget $f$ is the allowed fraction of the
additive KL of quantizing every layer, so $f=0.1$ allows $10\%$ of that accuracy loss. As in Section~\ref{sec:accbudget}, where bytes track
speed the two costs agree. With per-tensor FP8 on MI300A, FP8 is faster than BF16 even at batch~1, so both costs pick
nearly the same configurations, and on the Radeon~780M our cost ties the byte cost at 13 of 15 points.

A byte cost picks the same configurations on every device. Table~\ref{tab:xhw} shows what happens to the byte-cost
selections for Qwen2.5-7B when they move from MI300A to MI355X. On MI300A several of them are faster than BF16, but
on MI355X, where block-scaled FP8 is slower than BF16, every one of them is slower. Calibrated on the MI355X, our cost
instead keeps every layer in BF16 there.

\begin{table}[H]
\centering\small
\caption{Latency relative to BF16 of the same byte-priced Qwen2.5-7B configurations on MI300A $\rightarrow$ MI355X.
Values below 1 are faster than BF16.}
\label{tab:xhw}
\setlength{\tabcolsep}{3pt}
\begin{tabular}{lccc}\toprule
Byte-priced selection & Prefill & Decode $b{=}1$ & Decode $b{=}8$\\\midrule
$f=0.1$ & $0.88\rightarrow1.08$ & $1.05\rightarrow1.11$ & $0.98\rightarrow1.10$\\
$f=0.2$ & $0.80\rightarrow1.14$ & $1.05\rightarrow1.10$ & $0.97\rightarrow1.13$\\
$f=0.5$ & $0.70\rightarrow1.30$ & $1.66\rightarrow2.02$ & $1.02\rightarrow2.10$\\
All FP8 & $0.64\rightarrow1.31$ & $2.08\rightarrow2.57$ & $0.96\rightarrow2.65$\\\bottomrule
\end{tabular}
\end{table}

\subsection{Allocators derived from HAWQ-V3 and LLM-PQ}
\label{app:realalloc}
\begin{figure}[H]
\centering
\includegraphics[width=0.95\linewidth,height=0.45\textheight,keepaspectratio]{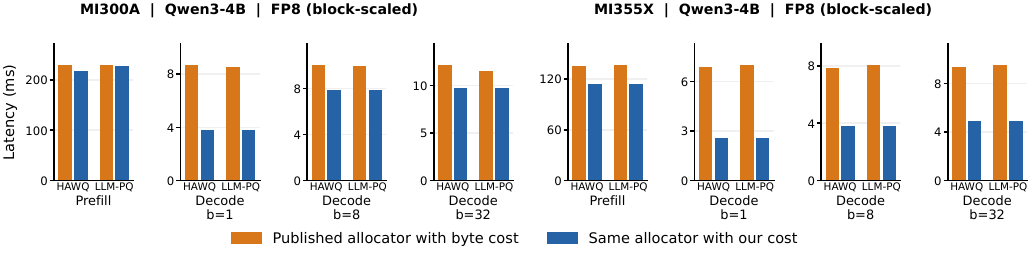}
\caption{Measured latency of the configurations that the HAWQ-V3- and LLM-PQ-derived allocators select with bytes or
with our cost (Qwen3-4B, BF16 or block-scaled FP8, largest budget).}
\label{fig:realalloc}
\end{figure}

Section~\ref{sec:accbudget} reports that with our cost the allocators derived from HAWQ-V3 and LLM-PQ select faster
configurations at 77 of 96 points. These are 44 of 48 on Qwen3-4B, 24 of 24 for Qwen2.5-7B on MI355X, and 9 of 24 on
MI300A, with ties otherwise. Figure~\ref{fig:realalloc} shows the measured latency of the Qwen3-4B selections at the largest budget.

\paragraph{Time costs built from measurements.}
\label{app:budgets}
Besides bytes, we compare our cost with two simpler time costs that also use measurements. The first times every
linear layer shape on its own with the stack's kernels and adds up these timings. For Qwen3-4B, it makes the same
selections as ours under accuracy budgets and is often faster than ours in decode under memory budgets. However, for
the MoE model it predicts about twice the time saving actually measured, because an expert layer timed on its own takes a
larger share of time than it does in a served step, so it misses speed targets that ours meets. The second is a simple
fix to the byte cost. We first run the whole model in each low-precision format, and if that is slower than BF16,
the allocator may not use the format at all. Otherwise it still prices time by bytes. Under accuracy budgets this
matches ours in most cases. But when a memory budget can only be met by using the slow format, the fix cannot avoid
it and falls back to plain bytes, and ours is faster in 33 of 48 cases on MI355X.

\paragraph{Speed budgets with MXFP4.} With BF16 or MXFP4 on Qwen3-4B (MI355X), our model mispredicts MXFP4 decode.
It is too optimistic at batch~1, where all-MXFP4 is slower than BF16, and too pessimistic at batches 8 and 32, where
it reports reachable targets as infeasible, so it meets 1 of 24 targets while bytes meet none. The batch-1 slowdown
comes from quantizing activations to 4 bits at every step, which costs more than it saves when a step has few tokens.
Switching to 16-bit activations at small batches removes it and lowers batch-1 decode time by about 25\% relative to
4-bit activations (Appendix~\ref{app:c3load}).

\subsection{After kernel tuning}
\label{app:kerneltuning}
Our cost model prices the kernels a stack actually runs, so when the kernels change, its selections should change
too. Here we tune the block-scaled FP8 kernels for the Qwen3-4B shapes on MI355X, also serve AITER's stock CK kernel
(Table~\ref{tab:kerneltuning}), and recalibrate.

\begin{table}[H]
\centering\small
\caption{All-FP8 latency relative to BF16 with four block-scaled FP8 kernels (Qwen3-4B, MI355X, 95\% interval over
four paired blocks).}
\label{tab:kerneltuning}
\setlength{\tabcolsep}{3pt}
\resizebox{0.7\linewidth}{!}{\begin{tabular}{lcccc}\toprule
FP8 kernel & Prefill & Decode $b{=}1$ & Decode $b{=}8$ & Decode $b{=}32$\\\midrule
vLLM Triton, shipped config & $1.24\pm.09$ & $2.78\pm.06$ & $2.22\pm.07$ & $1.75\pm.44$\\
vLLM Triton, tuned & $1.17\pm.04$ & $2.16\pm.04$ & $1.51\pm.06$ & $1.13\pm.31$\\
AITER CK, stock & $1.29\pm.07$ & $1.63\pm.04$ & $1.16\pm.04$ & $1.17\pm.37$\\
AITER Triton, tuned split-K & $1.21\pm.03$ & $1.03\pm.03$ & $0.81\pm.03$ & $0.87\pm.18$\\\bottomrule
\end{tabular}}
\end{table}

The first row is the shipped kernel our frozen model prices. With the tuned split-K kernel, FP8 decode is as fast
as BF16 or faster, while prefill stays slower with every kernel we tested.

\begin{figure}[H]
\centering
\includegraphics[width=0.68\linewidth,height=0.45\textheight,keepaspectratio]{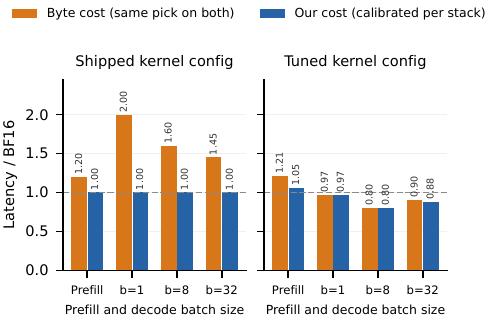}
\caption{Latency relative to BF16 of the byte-cost selections and ours, with the shipped and the tuned FP8 kernel
configuration (MI355X, Qwen3-4B, block-scaled FP8, same accuracy budget).}
\label{fig:tunedkernel}
\end{figure}

Recalibrated on the tuned stack with the same procedure, our model
moves from all-BF16 to FP8-heavy decode selections, while the byte-cost selections stay the same on both stacks
(Figure~\ref{fig:tunedkernel}).

\section{Batch size and load}
\label{sec:opdep-full}
\label{sec:actswitch-full}
This section gives the details behind Section~\ref{sec:opbatch}. It first asks whether the best weight configuration
changes with the operating point, then shows how the cost and accuracy of activation precision change with the number
of tokens and the phase, and how we switch activation precision at runtime.

\paragraph{Does the best weight configuration change?} Across 14 operating points for Qwen3-Coder-30B-A3B with
BF16/MXFP4, the best configuration changes, but one fixed configuration loses at most $0.9\%$ of modeled time, since
sensitivity decides which layers to quantize more than load does. At matched speed, per-point configurations have
$1.2$--$4.7\times$ lower additive KL at most points. Because switching weights costs memory or reload time, we switch
activation precision instead, which can change without touching stored weights.

\paragraph{Activation cost by token count.} Table~\ref{tab:moecost-full} shows how the relative cost of each format
moves with the number of tokens in one 256-expert, top-8 MoE layer on MI355X with AITER kernels~\citep{aiter}. Each
expert sees only a fraction of the tokens and stays weight-load bound, so extra activation arithmetic is nearly free
up to 128 tokens but doubles the layer time at prefill sizes.

\begin{table}[H]
\centering\small
\caption{Time per MoE layer in $\mu$s on MI355X (256 experts, top-8, hidden 7168, intermediate 2048). W4A4 uses MXFP4
weights and activations, W4\,A(4+1) adds one extra pass of activation arithmetic, and W4\,A16/A8 uses BF16
activations below 256 tokens and FP8 above. Percentages are relative to W4A4.}
\label{tab:moecost-full}
\setlength{\tabcolsep}{3pt}
\begin{tabular}{rrrrr}\toprule
Tokens & BF16 & W4A4 & W4\,A(4+1) & W4\,A16/A8\\\midrule
8 & 1027 & 267 & 273 (+2\%) & 282 (+5\%)\\
32 & 2886 & 735 & 747 (+2\%) & 769 (+5\%)\\
128 & 4484 & 1133 & 1167 (+3\%) & 1361 (+20\%)\\
512 & 4678 & 1232 & 1509 (+22\%) & 1223 ($-$1\%)\\
2048 & 5086 & 1552 & 1891 (+22\%) & 1674 (+8\%)\\
8192 & 8606 & 2817 & 5979 (+112\%) & 4111 (+46\%)\\\bottomrule
\end{tabular}
\end{table}

\begin{figure}[H]
\centering
\includegraphics[width=0.68\linewidth,height=0.45\textheight,keepaspectratio]{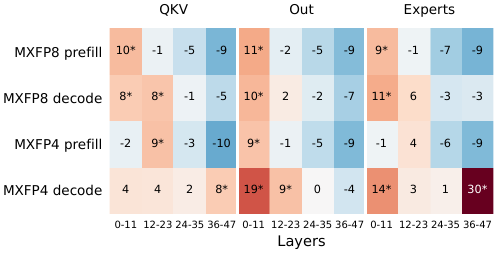}
\caption{Accuracy cost of lowering activation precision in one phase, summed per unit type and 12-layer band
(Qwen3-Coder-30B-A3B, MXFP4 weights). Values are the KL increase above the baseline from a change that barely
alters the activations, in units of $10^{-3}$, and $*$ marks sums more than two standard errors above that baseline.}
\label{fig:phasesens-full}
\end{figure}

\paragraph{Phase-specific accuracy costs.} We lower the activations of one unit at a time in Qwen3-Coder-30B-A3B,
on prompt rows only (prefill) or continuation rows only (decode), and measure KL to the BF16 model. In this MoE, even a
change that barely alters the activations shifts which experts some tokens are routed to and adds a little KL, so we
report each cost above this baseline (Figure~\ref{fig:phasesens-full}). MXFP4 activations cost almost no accuracy in prefill, where
they also save time. In decode they cost much more accuracy, while keeping activations at higher precision adds little
time at small batches, so higher-precision activations belong at small decode batches. In our serving runs, all
layers switch together by batch size. Most of the decode cost comes from a few layers, so choosing activation precision
per layer could keep only those layers at higher precision at larger batches as well.

\subsection{Under load}
\label{app:c3load}

\paragraph{Load sweep.} We add two formats to the cost model, MXFP4 weights with 16-bit or with 8-bit
activations, and fit their compute rates, bandwidths, and overheads from timing runs, as for the weight formats in
Section~\ref{sec:calib}. On configurations not used for fitting, the model's latency predictions are off by $5.4\%$
(Qwen3-4B), $6.2\%$ (Llama-3.1-8B), and $9.7\%$ (Qwen3-Coder-30B-A3B) on average. At each decode batch size, the
allocator then picks the activation format the model predicts is fastest. We send GSM8K requests with 256 output tokens at 1, 4, 16, and 64 concurrent requests
(Table~\ref{tab:c3load}). The fixed rule uses 16-bit activations whenever the decode kernels accept the step, up to
64 rows for the dense models, and for the MoE model it follows the AITER threshold, with 16-bit expert activations
below 8 tokens and 8-bit above. Apart from the MoE fixed rule, prefill runs 4-bit activations in every arm.
Table~\ref{tab:c3load} gives the numbers behind Figure~\ref{fig:c3}. Switching where our model says is up to 25\%
faster than 4-bit activations at low load and stays within about 1\% of them at 64 requests, while the fixed rule is
up to 23\% slower there. The predicted changes roughly track the measured ones.

\begin{table}[H]
\centering\small
\caption{Time per output token with activation switching, relative to always using 4-bit activations (MI355X, MXFP4
weights). The 4-bit column gives this baseline in ms, and the other columns give the change from it, where negative is
faster. Brackets give our model's predicted change. Intervals are 95\% over three paired blocks, or five for
Llama-3.1-8B and the MoE model.}
\label{tab:c3load}
\setlength{\tabcolsep}{3pt}
\resizebox{0.7\linewidth}{!}{\begin{tabular}{llrrr}\toprule
Model & Requests & 4-bit (ms) & Fixed rule (\% change) & Cost-model policy (\% change)\\\midrule
Qwen3-4B & 1 & 2.64 & $-25.1\pm0.7\%$ & $-24.9\pm1.0\%$ [$-22.1\%$]\\
 & 4 & 2.84 & $-18.3\pm1.9\%$ & $-21.4\pm0.9\%$ [$-18.6\%$]\\
 & 16 & 2.87 & $+9.5\pm0.7\%$ & $-4.5\pm1.7\%$ [$-10.3\%$]\\
 & 64 & 4.02 & $+22.8\pm3.5\%$ & $-0.1\pm4.7\%$ [$+0.0\%$]\\\midrule
Llama-3.1-8B & 1 & 2.65 & $-13.3\pm0.7\%$ & $-11.1\pm1.4\%$ [$-13.4\%$]\\
 & 4 & 2.82 & $-12.7\pm1.7\%$ & $-11.9\pm2.0\%$ [$-10.4\%$]\\
 & 16 & 3.01 & $-3.8\pm2.2\%$ & $-2.6\pm3.1\%$ [$-1.4\%$]\\
 & 64 & 4.37 & $+10.1\pm2.2\%$ & $+0.9\pm0.7\%$ [$+0.0\%$]\\\midrule
Qwen3-Coder-30B & 1 & 3.85 & $-8.0\pm1.3\%$ & $-7.9\pm0.6\%$ [$-2.7\%$]\\
 & 4 & 4.19 & $-5.1\pm1.1\%$ & $-5.0\pm0.9\%$ [$-9.0\%$]\\
 & 16 & 4.89 & $+0.7\pm1.5\%$ & $-0.3\pm1.4\%$ [$+0.0\%$]\\
 & 64 & 6.21 & $+1.4\pm1.5\%$ & $-0.8\pm1.8\%$ [$+0.0\%$]\\
\bottomrule
\end{tabular}}
\end{table}

\paragraph{Accuracy at batch 1.} We evaluate one request at a time, so every decode step uses the batch-1 format. On
the full GSM8K test set, the policy raises accuracy over 4-bit activations by $+3.6\pm1.8$ points on Qwen3-4B and
$+2.6\pm2.1$ points on Llama-3.1-8B. Decode KL to BF16 falls by $49\%$ on Qwen3-4B, $39\%$ on Llama-3.1-8B, and $25\%$
on Qwen3-Coder-30B-A3B.

\subsection{With arriving requests}
\label{app:openloop}

We repeat the load sweep with Poisson arrivals at four rates on a vLLM server with chunked prefill. At the lowest
rate, the policy lowers median time per output token by $25.6\%$ (Qwen3-4B) and $12.8\%$ (Llama-3.1-8B) relative to
4-bit activations. At the highest rate, the fixed rule is $17.3\pm7.3\%$ and $5.2\pm0.3\%$ slower than 4-bit
activations, while the policy ties. For Qwen3-Coder-30B-A3B, the policy saves $6.0\%$ at the lowest rate but costs
about $1\%$ at high load because of the prototype's dispatch overhead.

\section{Prefill and decode}
\label{app:disagg}
Prefill and decode stress different resources and differ in accuracy sensitivity. This section gives the per-phase
results behind Section~\ref{sec:opphase}.

\subsection{Separate GPUs for each phase}

For each budget, our cost model selects one configuration for prefill and one for decode. On MI300A with
block-scaled FP8, the decode selection is BF16 at every batch size. We compare this pair against BF16, our prefill
selection, and the byte-cost selection, each deployed for both phases (Table~\ref{tab:disagg}). Latency excludes the
KV-cache transfer, and for accuracy we run the prompt under the prefill configuration and decode under the decode
configuration. The pair cuts decode KL by $73$--$76\%$ relative to using the prefill selection for both phases. On
MI355X with MXFP4, it is within about $\pm2\%$ of the best single deployment.

\paragraph{Real disaggregation with per-phase tables (MI355X, MXFP4).} We serve Qwen3-4B with prefill and decode on
separate MI355X GPUs. Per-unit KL measured at prompt positions and at generated positions gives prefill and decode
tables whose rankings barely agree (Spearman $0.15$). With per-phase tables, the pair has the lowest decode KL of all
arms ($0.062$, against $0.069$ for one shared table and $0.071$ for bytes) at the same time per output token.

\begin{table}[H]
\centering\small
\caption{Disaggregated prefill and decode on MI300A with block-scaled FP8. The accuracy budget $f$ is the allowed
fraction of the additive KL of quantizing every layer. Accuracy is in percent, decode KL to BF16 is in units of
$10^{-3}$, and time to first token (TTFT) and time per output token (TPOT, batch~1) are in ms.}
\label{tab:disagg}
\setlength{\tabcolsep}{1.5pt}
\resizebox{0.8\linewidth}{!}{\begin{tabular}{llccccc}\toprule
Model, budget & Arm & TTFT & TPOT & GSM8K & MATH & KL\\\midrule
Qwen3-4B, $f=0.5$ & BF16 both & 289 & 3.8 & 94.7 & 71.0 & 0\\
 & Byte both & 243 & 6.6 & 95.0 & 67.0 & 9.4\\
 & Prefill sel.\ both & 224 & 4.9 & 93.7 & 70.5 & 9.3\\
 & Per-phase & 224 & 3.8 & 94.3 & 70.0 & 2.5\\\midrule
Qwen3-4B, $f=0.2$ & Prefill sel.\ both & & & 93.3 & 68.5 & 5.3\\
 & Per-phase & & & 93.3 & 69.5 & 1.4\\\midrule
Qwen2.5-7B, $f=0.5$ & BF16 both & & & 93.0 & 67.0 & 0\\
 & Prefill sel.\ both & & & 92.0 & 69.0 & 5.4\\
 & Per-phase & & & 92.7 & 69.5 & 1.3\\\bottomrule
\end{tabular}}
\end{table}

\subsection{Shared GPUs}
\label{app:colocated}

\paragraph{Predicting mixed steps.} With chunked prefill, a step can contain prompt chunks $c$ of $s_c$ tokens, each
following $\ell_c$ cached tokens, and decode tokens $i$ with context lengths $L_i$. This gives $n_p=\sum_c s_c$
prompt tokens and $n_d$ decode tokens. Every weight multiplies all $n_p+n_d$ rows, so the unit terms keep their form,
and we sum the baseline and attention terms over what the step contains:
\begin{equation}
\begin{aligned}
T_\text{step} &= f_\text{step} + f_\text{prompt}\,[n_p>0] + \beta_p n_p + \beta_d n_d\\
&\quad + a_\text{prefill}\textstyle\sum_c s_c(2\ell_c+s_c) + a_\text{decode}\sum_i L_i\\
&\quad + \textstyle\sum_u T_\text{unit}\big(u,\text{cfg}[u];\,n_p+n_d\big).
\end{aligned}
\label{eq:mixed}
\end{equation}
Pure prefill and pure decode recover the $BS^2$ and $BL$ attention terms of Section~\ref{sec:method}. Fitted on
Qwen3-4B (MI355X) and frozen, this model has $7.3\%$ mean error on 612 held-out mixed steps, against $14.7\%$ for
adding separate prefill and decode predictions and $8.5\%$ for a linear proxy on tokens and bytes.

\paragraph{Choosing weights for a measured step mix.} Qwen3-4B serving GSM8K traffic at 16.8 requests/s is about
$95\%$ pure decode. With costs weighted by this mix, the allocator returns the decode selection, which serves at
$2.92$\,ms per output token, against $3.33$\,ms for the prefill selection and $3.61$\,ms for BF16.

\section{Context length}
\label{app:kvcontext}
As context grows, KV-cache reads take a larger share of each decode step. This section covers how well the model
predicts this, cache precision, and the schedules that lower weight precision as context grows.

\subsection{Decode latency as the KV cache grows}

\begin{figure}[H]
\centering
\includegraphics[width=0.95\linewidth,height=0.45\textheight,keepaspectratio]{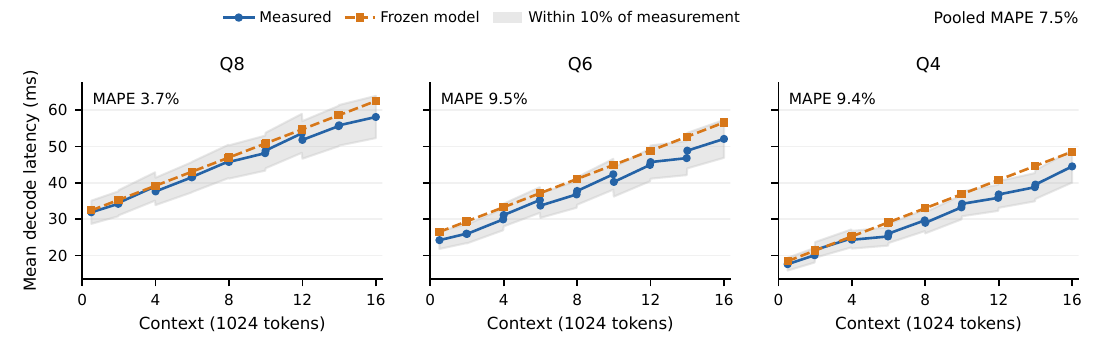}
\caption{Decode latency as context grows to 16K, with fixed Q8/Q6/Q4 weights and an F16 KV cache. Squares are
frozen predictions and gray bands show $\pm10\%$ of the measurements.}
\label{fig:kvcontext}
\end{figure}

We fit the cost model, with one bandwidth shared by all formats, on Llama-3.2-1B-Instruct~\citep{llama32} on the
Radeon~780M (llama.cpp, batch~1, F16 KV cache) over contexts from 512 to 16384 tokens, and predict fixed Q8, Q6, and
Q4 decode on held-out passages (Figure~\ref{fig:kvcontext}). Errors are $3.7\%$, $9.5\%$, and $9.4\%$. The same
effect appears on datacenter GPUs. On MI355X with Qwen3-4B, FP8 decode takes $2.24\times$ the BF16 step time at
batch~32 and context 512 but $1.40\times$ at 16K tokens ($2.12\times$ to $1.76\times$ at batch~8), and MI300A shows
the same trend at batch~8 ($1.31\times$ to $1.14\times$). The weight format matters less as reading the KV cache
takes a larger share of each step.

\subsection{Cache precision}
\label{app:kvdegrad}

\paragraph{Enumerating cache precisions.}
For each cache format $\kappa$ at context length $L$, the time left for weight operations is
\[
D_\mathrm{weights}(\kappa,L)=D-f_\text{base}-T_\text{attention}(\kappa,L).
\]
For each $\kappa$ we compute $T_0$, run Algorithm~\ref{alg:allocate} with $\tau=D/T_0-1$, and discard
infeasible pairs. Every remaining pair of cache format and weight configuration meets the latency budget, so choosing
among them means comparing the error from the cache with the error from the weights, as below.

\paragraph{How does a quantized KV cache affect the accuracy budget?}
We model the cache's accuracy cost as one more term added to the weight costs,
\begin{equation}
\widehat{A}_\text{total}=\sum_u A\big(u,\text{cfg}[u]\big)+A_\text{kv}(\kappa,L),
\label{eq:kltotal}
\end{equation}
where $A_\text{kv}$ is measured with reference weights.

\subsection{Running the schedule}
\label{sec:hardware}

We run Q8/Q6/Q4 weights with an F16 cache on the Radeon~780M, using the calibration from
Appendix~\ref{app:kvcontext}. For scheduling we add a paired term to the calibration objective that fits
differences between each control and its adjacent reference runs:
\begin{equation}
\begin{aligned}
J_{\mathrm{schedule}}&=J_{\mathrm{base}}+\tfrac12 C_{\mathrm{pair}},\\
C_{\mathrm{pair}}&=\frac{1}{n}\sum_{j=1}^n
\left(\frac{\widehat T_{j,\mathrm{control}}-\widehat T_{j,\mathrm{reference}}-\delta_j}{r_j}\right)^2.
\end{aligned}
\label{eq:scheduleloss}
\end{equation}
Here $\delta_j$ is the measured control-minus-reference latency and $r_j$ is the mean reference latency. The
byte-priced baseline uses the same objective. Both allocators reserve a $4.04$\,ms timing margin and a $2.5$\,ms
switching allowance.

\paragraph{Evaluation.}
\label{app:scheduleconfirm}
A modified llama.cpp keeps Q8, Q6, and Q4 copies of every weight and switches at scheduled context lengths. We
compare cost-model and byte-priced schedules for a 40\,ms target with fixed Q8, Q6, and Q4 on six held-out passages
through 16K tokens (Table~\ref{tab:schedulererun}).

\begin{table}[H]
\centering\small
\caption{Context-length schedules on the Radeon~780M (Llama-3.2-1B, six passages, two processes, 4{,}608 timed
steps per policy).}
\label{tab:schedulererun}
\setlength{\tabcolsep}{3pt}
\resizebox{0.7\linewidth}{!}{\begin{tabular}{lrrrrr}\toprule
Policy & Mean (ms) & P90 (ms) & Steps $>60$\,ms & KL to Q8 ($10^{-3}$) & Perplexity\\\midrule
Cost-model schedule & 38.4 & 53.3 & 13 & 6.6 & 11.17\\
Byte-priced schedule & 39.4 & 54.7 & 7 & 10.1 & 11.20\\
Fixed Q8 & 39.8 & 54.9 & 122 & 0 & 11.05\\
Fixed Q6 & 41.7 & 57.6 & 314 & 6.1 & 11.06\\
Fixed Q4 & 33.9 & 48.8 & 4 & 80.2 & 12.23\\\bottomrule
\end{tabular}}
\end{table}

Both schedules keep nearly every step within 60\,ms, while fixed Q8 and Q6 exceed it on 122 and 314 steps. The
cost-model schedule is $2.7\%$ faster than the byte-priced schedule and has $35\%$ lower KL. Laptop speed drifts
by about $30\%$, so no policy meets 40\,ms on every step.

\subsection{Tool use}
\label{sec:slmruntime-full}

\begin{figure}[H]
\centering
\includegraphics[width=0.68\linewidth,height=0.45\textheight,keepaspectratio]{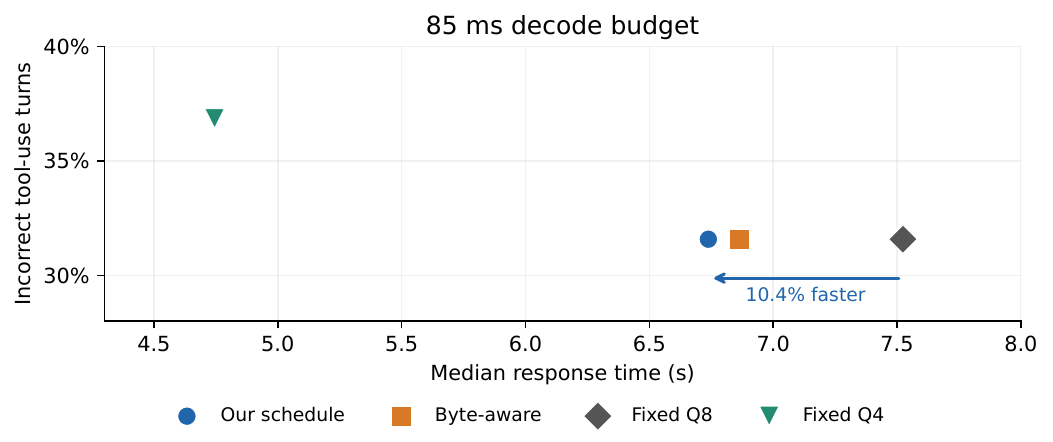}
\caption{Tool-use error versus median response time at an 85\,ms decode budget (Qwen2.5-3B-Instruct,
Radeon~780M).}
\label{fig:slmagents}
\end{figure}

We run six held-out BFCL multi-turn tasks~\citep{bfcl} (19 turns) with
Qwen2.5-3B-Instruct~\citep{qwen25three} at an 85\,ms decode budget and score actions with the official
simulator. Figure~\ref{fig:slmagents} plots error against median response time. Our schedule answers the same
13 turns correctly as Q8 with lower response time ($6.74$\,s versus $7.52$\,s).

\subsection{Switching precision without resident copies}
\label{app:noresident}
Our runtime keeps every precision in memory, so a switch only changes which copy a tensor uses. Because a schedule
is fixed before decoding, the runtime knows which tensor versions it needs long before it needs them, so the same
schedules can also run without resident copies~\citep{athmaram2026contextaware}. Each tensor version can be streamed
from storage before its switch, quantized in place from the most precise version in memory, or stored as a 4-bit
base plus a residual that is turned on or off. Streaming gives outputs
identical to the resident runtime and lowers peak weight memory from 3.23\,GB to 2.35\,GB on Llama-3.2-1B-Instruct,
with about 0.07\,ms of loading per decode step.

\section{Accuracy measurements}
\label{sec:analysis}
The allocator's accuracy cost comes from measuring one unit at a time and adding the results. This section shows how
well the sum tracks the joint error and the accuracy table for each model.

\subsection{How well additive KL tracks joint KL}
\label{sec:opdivergence}

\begin{figure}[H]
\centering
\includegraphics[width=0.95\linewidth,height=0.78\textheight,keepaspectratio]{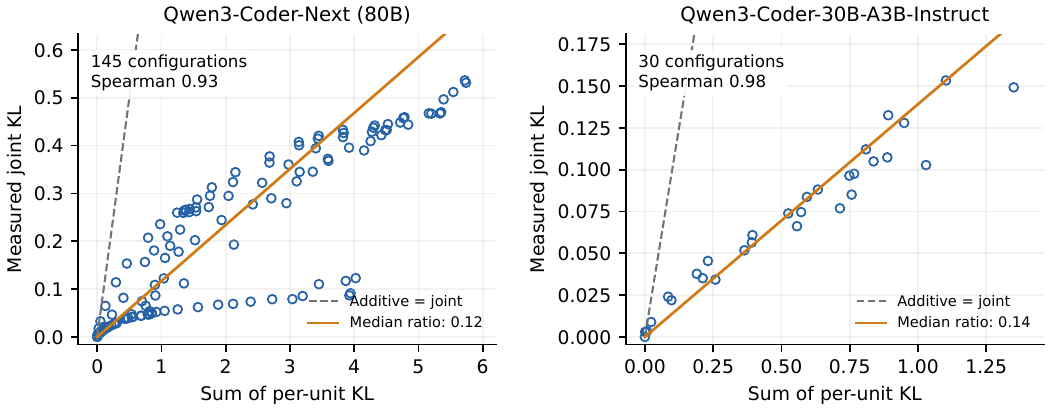}
\caption{Additive versus joint KL for Qwen3-Coder-Next~\citep{qwen3codernext} and
Qwen3-Coder-30B-A3B-Instruct~\citep{qwen3coder30b} configurations.}
\label{fig:calibration}
\end{figure}

Figure~\ref{fig:calibration} compares the additive accuracy cost with measured joint KL for two models. The sum
ranks configurations well (Spearman $0.93$ and $0.98$) but overestimates joint KL, and it is not a guaranteed
bound. Selected allocations are therefore evaluated directly.

\subsection{Accuracy cost of each unit}
\label{app:acctables}
\begin{figure}[H]
\centering
\includegraphics[width=0.95\linewidth]{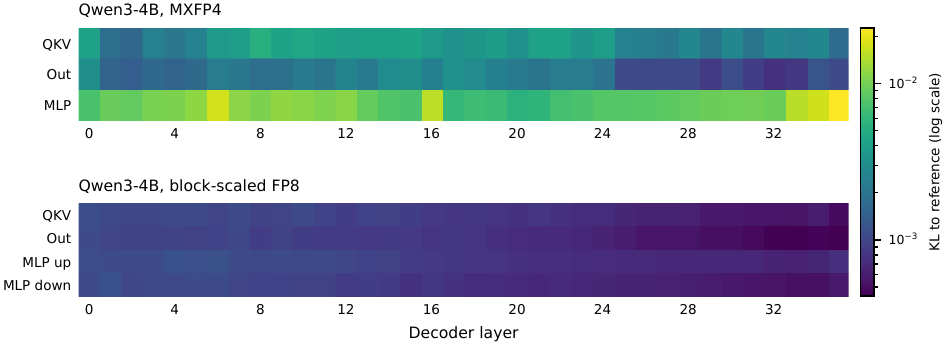}
\caption{Accuracy cost per unit for Qwen3-4B. KL to BF16 when one unit is quantized.}
\label{fig:acc_q3}
\end{figure}

\begin{figure}[H]
\centering
\includegraphics[width=0.95\linewidth]{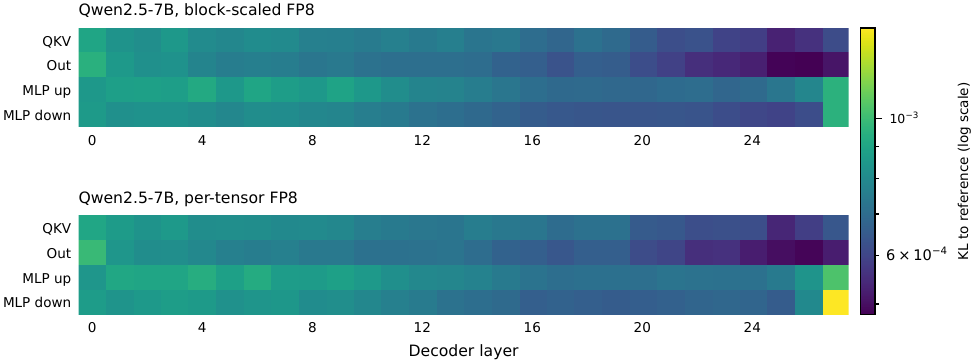}
\caption{Accuracy cost per unit for Qwen2.5-7B. KL to BF16 when one unit is quantized.}
\label{fig:acc_q25}
\end{figure}

\begin{figure}[H]
\centering
\includegraphics[width=0.95\linewidth]{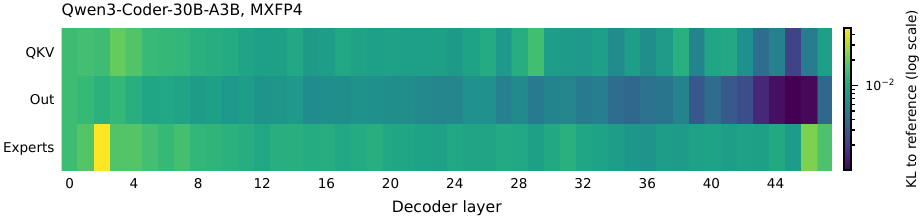}
\caption{Accuracy cost per unit for Qwen3-Coder-30B-A3B. KL to BF16 when one unit is quantized.}
\label{fig:acc_coder}
\end{figure}

\begin{figure}[H]
\centering
\includegraphics[width=0.95\linewidth]{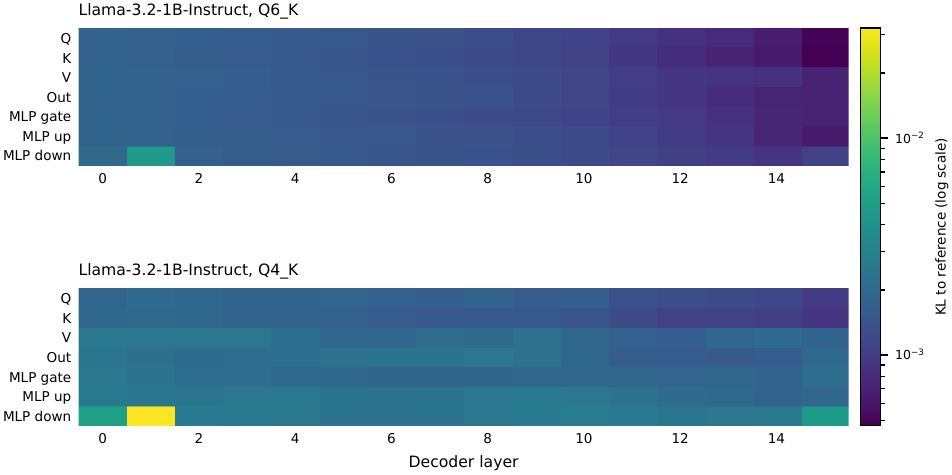}
\caption{Accuracy cost per tensor for Llama-3.2-1B-Instruct. KL to F16 when one tensor is quantized.}
\label{fig:acc_llama}
\end{figure}

Figures~\ref{fig:acc_q3}--\ref{fig:acc_llama} show the accuracy tables for the models in our experiments. Each value
is the KL to the reference model when only that unit is quantized. Each serving experiment uses the table measured
for its own model and format.

\section{Mixed-precision configurations behind the main results}
\label{app:configs}
This section collects the per-layer configurations behind the main-text figures, so that each result can be traced
to the configuration that produced it. In every heatmap, blue is the reference precision and orange the lower
precision.

\subsection{Figure~\ref{fig:teaser}}
\label{app:fig1configs}

\begin{figure}[H]
\centering
\includegraphics[width=0.95\linewidth,height=0.45\textheight,keepaspectratio]{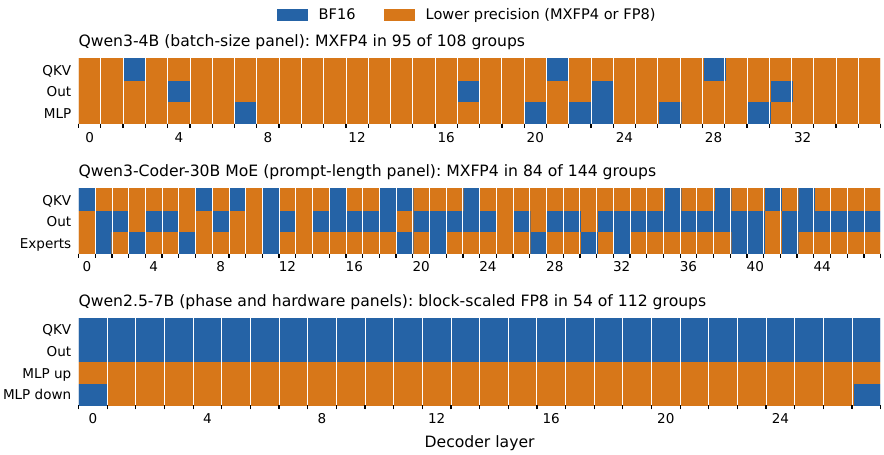}
\caption{Configurations for Figure~\ref{fig:teaser}: per-layer formats of the lower-precision configuration in each panel.
The Qwen3-4B and Qwen3-Coder-30B configurations come from Section~\ref{sec:predict}, the Qwen2.5-7B configuration is the
byte-based selection at the largest budget in Section~\ref{sec:swap}, and the context-length panel uses uniform Q4
against uniform Q8 with an F16 KV cache.}
\label{fig:fig1configs}
\end{figure}

\subsection{Memory budget (Figure~\ref{fig:membudget})}

\begin{figure}[H]
\centering
\includegraphics[width=0.95\linewidth,height=0.8\textheight,keepaspectratio]{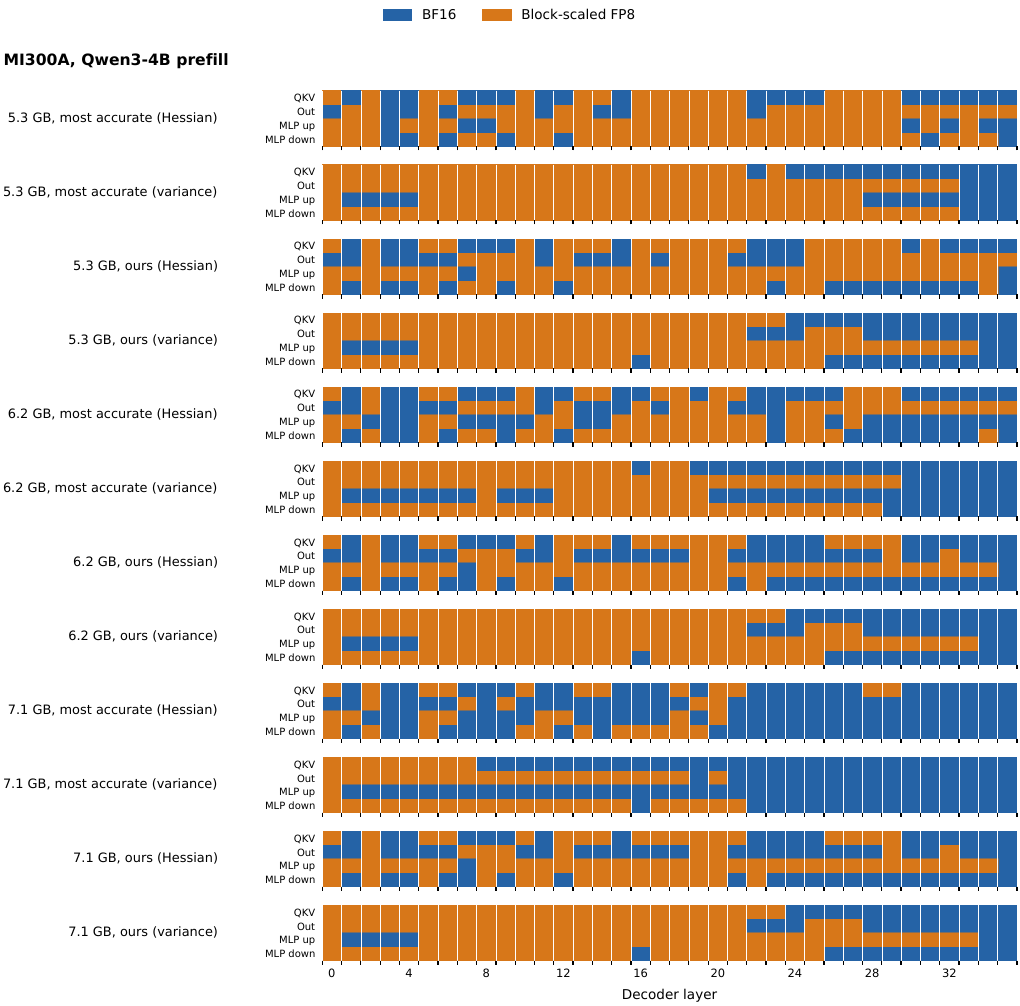}
\caption{Configurations for Figure~\ref{fig:membudget}, MI300A: prefill selections of the most accurate configuration that fits and of our cost at each memory limit (Qwen3-4B, BF16 or block-scaled FP8), with the Hessian-trace and variance sensitivity measures.}
\label{fig:cfgmem300}
\end{figure}

\begin{figure}[H]
\centering
\includegraphics[width=0.95\linewidth,height=0.8\textheight,keepaspectratio]{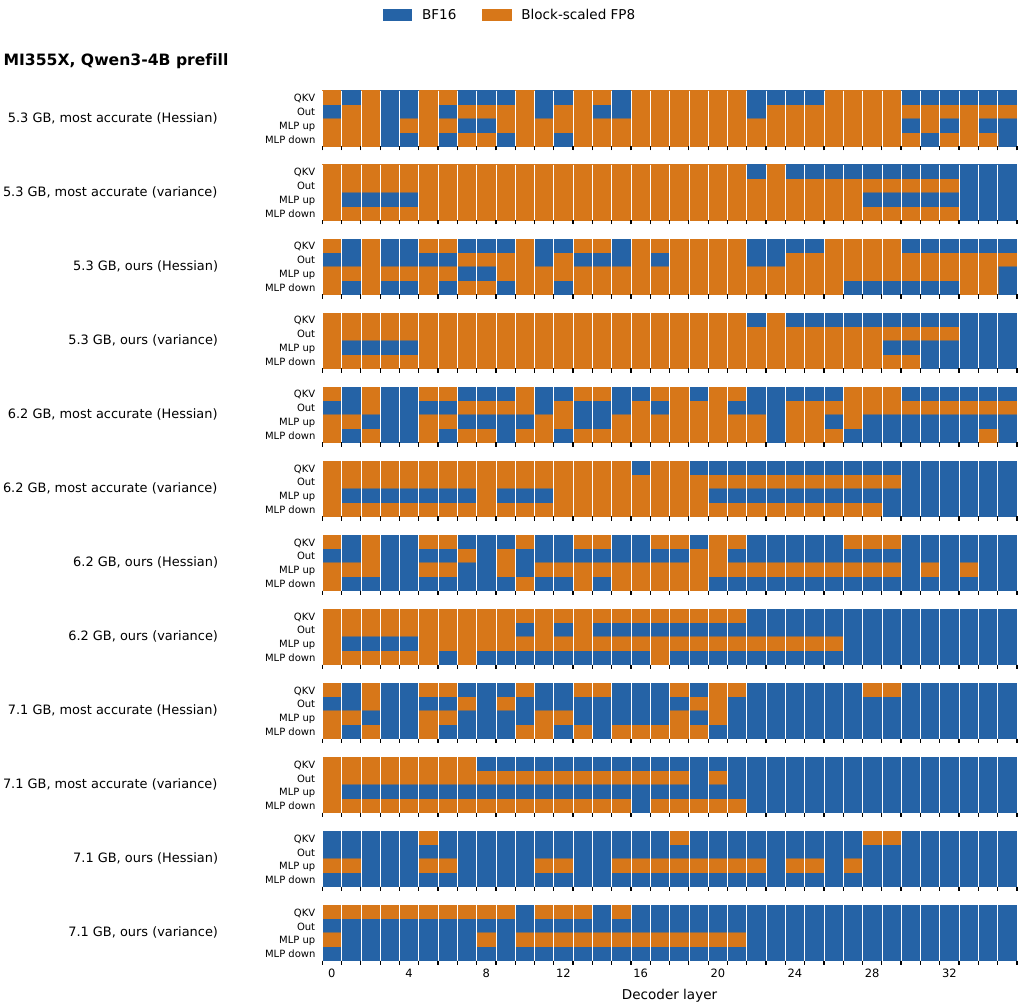}
\caption{Configurations for Figure~\ref{fig:membudget}, MI355X: prefill selections, as in Figure~\ref{fig:cfgmem300}.}
\label{fig:cfgmem355}
\end{figure}

\subsection{Speed budget (Figure~\ref{fig:speedbudget})}
With our cost, every decode selection under a speed budget is BF16, so only prefill on MI300A and the MoE model are
shown. For the MoE model, every selection keeps QKV and attention output in BF16, and missing rows are targets that our
cost reported as infeasible.

\begin{figure}[H]
\centering
\includegraphics[width=0.95\linewidth,height=0.8\textheight,keepaspectratio]{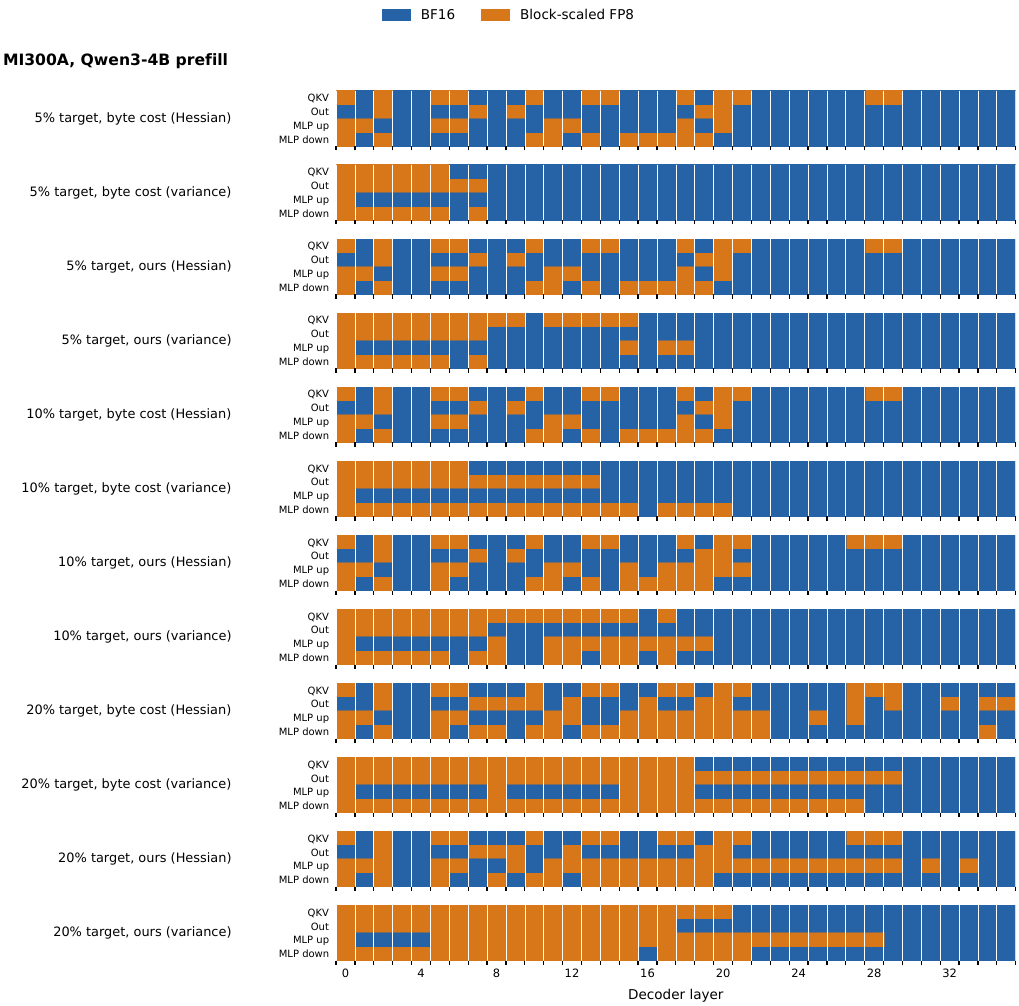}
\caption{Configurations for Figure~\ref{fig:speedbudget}, middle: MI300A prefill selections at each requested latency reduction with a byte cost and with our cost (Qwen3-4B, BF16 or block-scaled FP8).}
\label{fig:cfgspeed}
\end{figure}

\begin{figure}[H]
\centering
\includegraphics[width=0.95\linewidth,height=0.8\textheight,keepaspectratio]{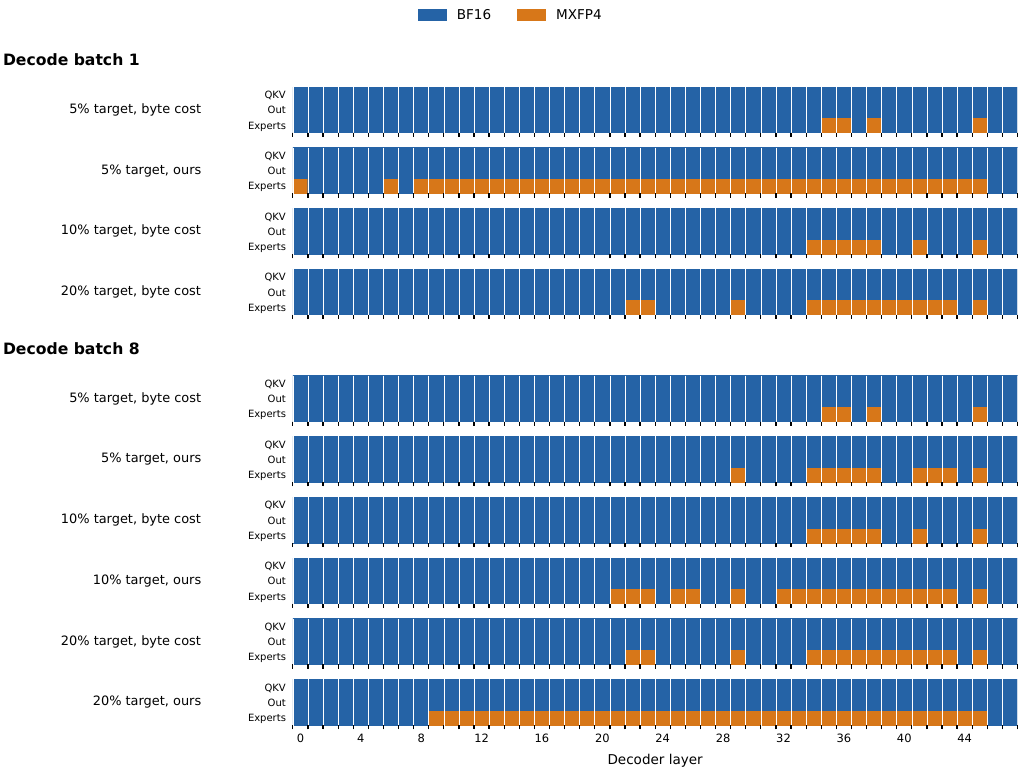}
\caption{Configurations for Figure~\ref{fig:speedbudget}, right, at decode batch sizes 1 and 8: selections at each requested latency reduction with a byte cost and with our cost (Qwen3-Coder-30B-A3B, MI355X, BF16 or MXFP4).}
\label{fig:cfgspeedmoe}
\end{figure}

\begin{figure}[H]
\centering
\includegraphics[width=0.95\linewidth,height=0.8\textheight,keepaspectratio]{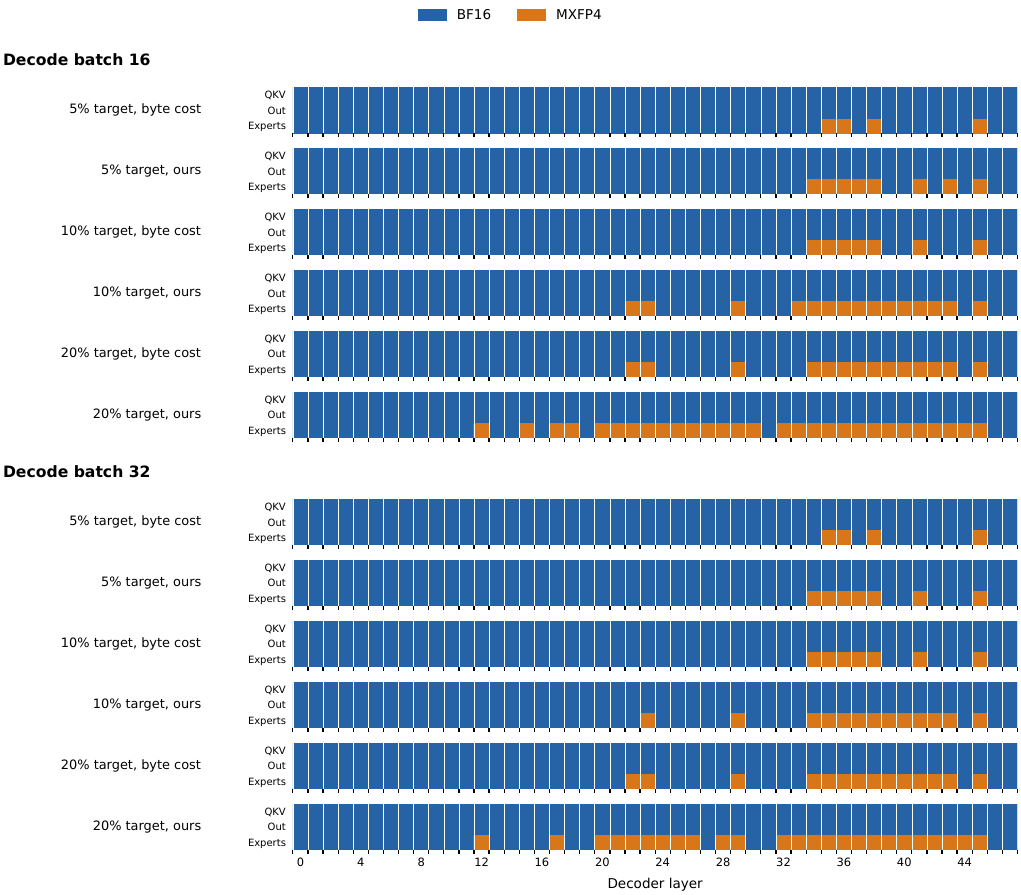}
\caption{Configurations for Figure~\ref{fig:speedbudget}, right, at decode batch sizes 16 and 32, as in Figure~\ref{fig:cfgspeedmoe}.}
\label{fig:cfgspeedmoeb}
\end{figure}

\subsection{Accuracy budget (Figure~\ref{fig:c2bars})}
These are the selections at the largest budget. The MI300A selections are also the ones deployed per phase in
Figure~\ref{fig:disagg}, with our prefill selection on the prefill pool and BF16 on the decode pool.

\begin{figure}[H]
\centering
\includegraphics[width=0.95\linewidth,height=0.8\textheight,keepaspectratio]{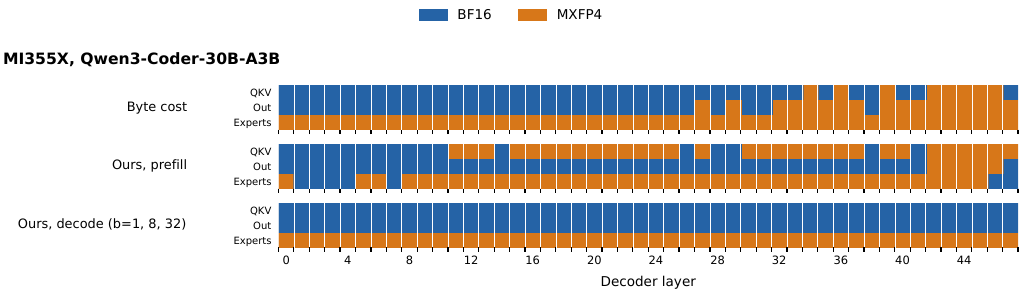}
\caption{Configurations for Figure~\ref{fig:c2bars}, left: selections with a byte cost and with our cost at the largest accuracy budget (Qwen3-Coder-30B-A3B, MI355X, BF16 or MXFP4).}
\label{fig:cfgaccmoe}
\end{figure}

\begin{figure}[H]
\centering
\includegraphics[width=0.95\linewidth,height=0.8\textheight,keepaspectratio]{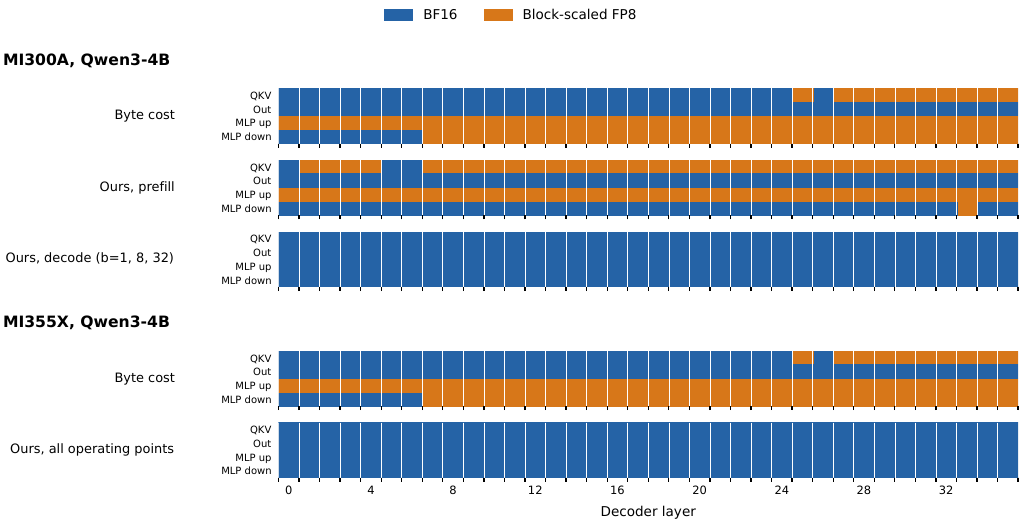}
\caption{Configurations for Figure~\ref{fig:c2bars}, middle and right: selections with a byte cost and with our cost at the largest accuracy budget (Qwen3-4B, BF16 or block-scaled FP8). The MI300A selections are also those deployed per phase in Figure~\ref{fig:disagg}.}
\label{fig:cfgaccq3}
\end{figure}

\subsection{Hardware (Figure~\ref{fig:configsxhw})}
Figure~\ref{fig:configsxhw} serves the memory-budget prefill selections at 6.2 and 7.1\,GB on both devices and averages
over the two limits and sensitivity measures. The configuration chosen for MI300A uses FP8 widely, while the one
chosen for MI355X keeps more layers in BF16.

\begin{figure}[H]
\centering
\includegraphics[width=0.95\linewidth,height=0.8\textheight,keepaspectratio]{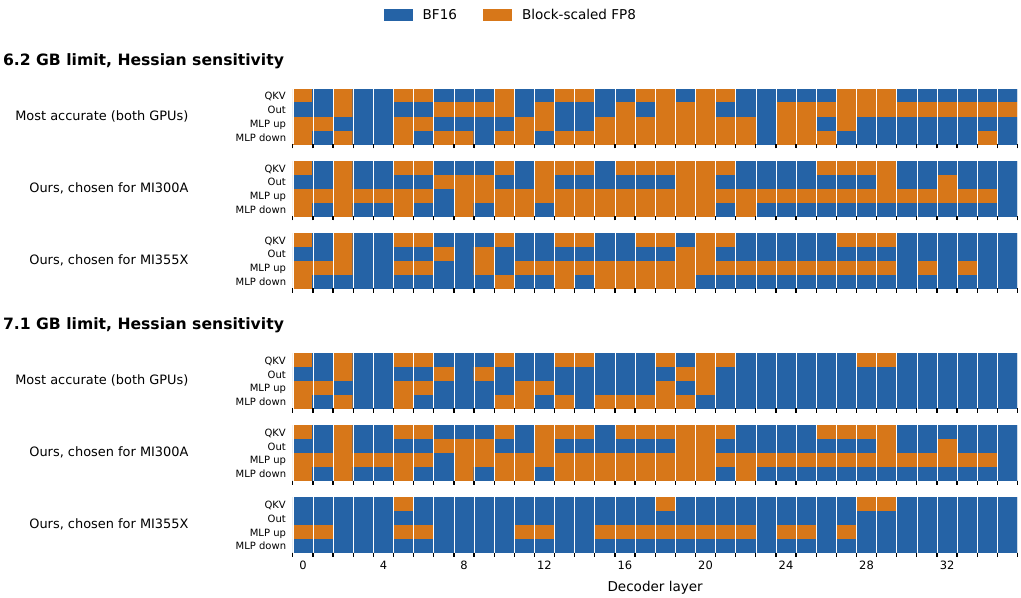}
\caption{Configurations for Figure~\ref{fig:configsxhw} with the Hessian-trace sensitivity: the most accurate configuration that fits, which is the same on both GPUs, and the selections with our cost chosen for each GPU (Qwen3-4B prefill, BF16 or block-scaled FP8). The accuracy costs are the same on both GPUs, so the selections differ only because of each GPU's calibrated time costs.}
\label{fig:cfghw1}
\end{figure}

\begin{figure}[H]
\centering
\includegraphics[width=0.95\linewidth,height=0.8\textheight,keepaspectratio]{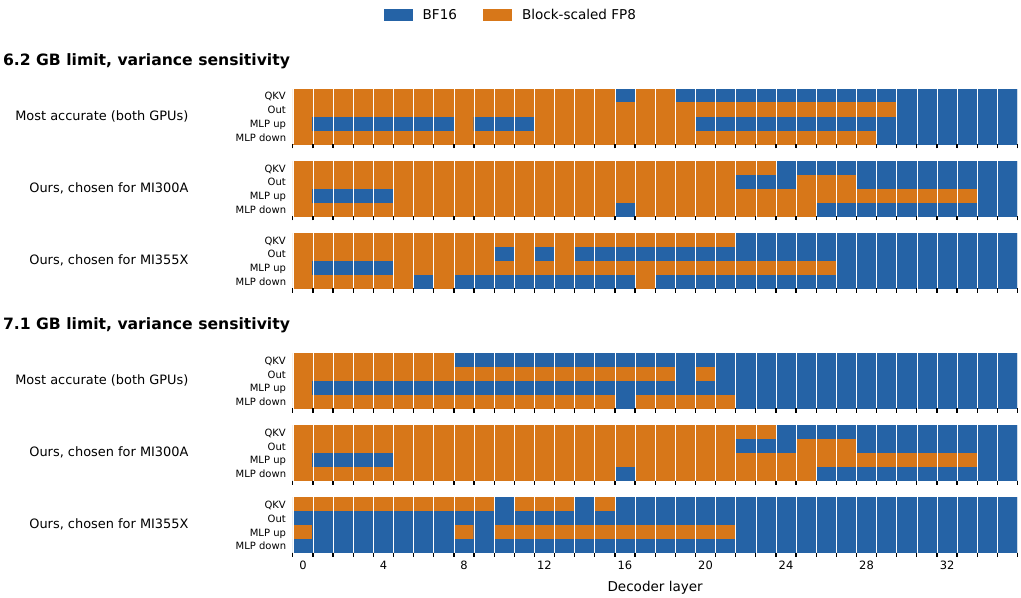}
\caption{Configurations for Figure~\ref{fig:configsxhw} with the variance sensitivity, as in Figure~\ref{fig:cfghw1}.}
\label{fig:cfghw2}
\end{figure}

\section*{AI Usage Disclosure}
We used LLM-based coding agents (Codex) throughout this work under the authors' direction. They helped search
and summarize related work, propose experiment protocols, write and run experiment and analysis code, and draft and
edit text. The research question, the cost model, and all framing decisions are the authors'. The authors reviewed
all code, results, and text, checked every citation against its source, and take full responsibility for the paper.
Every reported number comes from measurements on the hardware described.

\end{document}